\documentclass[sigconf]{acmart}

\usepackage{graphicx}        % \scalebox
\usepackage{booktabs}        % \toprule \midrule \bottomrule
\usepackage{makecell}        % \makecell
\usepackage[table]{xcolor}   % \rowcolor and gray!10
\usepackage{caption}         % \captionsetup
\usepackage{natbib}          % \citep  (if you use \citep)

\usepackage{algorithm}

\usepackage{adjustbox} % important: scales DOWN only if needed

\AtBeginDocument{%
  }

\copyrightyear{2026}
\acmYear{2026}
\setcopyright{cc}
\setcctype{by}
\acmConference[KDD '26]{Proceedings of the 32nd ACM SIGKDD Conference on Knowledge Discovery and Data Mining V.2}{August 09--13, 2026}{Jeju Island, Republic of Korea}
\acmBooktitle{Proceedings of the 32nd ACM SIGKDD Conference on Knowledge Discovery and Data Mining V.2 (KDD '26), August 09--13, 2026, Jeju Island, Republic of Korea}
\acmDOI{10.1145/3770855.3817885}
\acmISBN{979-8-4007-2259-2/2026/08}

\usepackage[dvipsnames, table]{xcolor}
\usepackage{pifont}

\usepackage{booktabs}
\usepackage{multirow}
\usepackage[table]{xcolor}
\usepackage{graphicx}
\usepackage{pifont}

\usepackage{xcolor}

\usepackage{multirow}
\usepackage{enumitem}
\usepackage[table]{xcolor} % Required for coloring table cells
\usepackage[table]{xcolor} % Required for coloring table cells
\usepackage{graphicx}

\usepackage{amsmath}
\usepackage{booktabs}
\usepackage{algorithm}
\usepackage{algorithmic}
\begin{document}
%\title{From Effectiveness to Adversarial Robustness: Advancing Multimodal Medical Image Analysis}
\title{Advancing Multimodal Fusion on Heterogeneous Medical Data with Hybrid Geometry Attention}


% KDD / ACM acmart author block
% Use with: \documentclass[sigconf]{acmart}

% KDD / ACM acmart camera-ready author block
% Use with: \documentclass[sigconf]{acmart}
% Do not wrap the author block in \small; acmart controls the camera-ready font.
% Optional compact layout, supported by acmart. Use only if the author block is too tall.
% \settopmatter{authorsperrow=3}

\author{Joy Dhar}
\orcid{0000-0001-9488-8298}
\affiliation{%
  \institution{Indian Institute of Technology Ropar}
  \city{Rupnagar}
  \state{Punjab}
  \country{India}
}
\email{joy.22csz0003@iitrpr.ac.in}

\author{Manish Kumar Pandey}
\orcid{0009-0001-1619-5741}
\affiliation{%
  \institution{RoentGen Health}
  \city{New Delhi}
  \state{Delhi}
  \country{India}
}
\email{manishpandey@roentgenhealth.com}

\author{Nayyar Zaidi}
\orcid{0000-0003-4024-2517}
\affiliation{%
  \institution{Deakin University}
  \city{Melbourne}
  \state{Victoria}
  \country{Australia}
}
\email{nayyar.zaidi@deakin.edu.au}

\author{Chen Chen}
\orcid{0000-0003-3957-7061}
\affiliation{%
  \institution{University of Central Florida}
  \city{Orlando}
  \state{Florida}
  \country{USA}
}
 \email{chen.chen@crcv.ucf.edu}

\author{Maryam Haghighat}
\orcid{0000-0002-2080-8483}
\affiliation{%
  \institution{Queensland University of Technology}
  \city{Brisbane}
  \state{QLD}
  \country{Australia}
}
\email{maryam.haghighat@qut.edu.au}

\author{Ferdous Sohel}
\orcid{0000-0003-1557-4907}
\affiliation{%
  \institution{Murdoch University}
  \city{Perth}
  \state{Western Australia}
  \country{Australia}
}
 \email{f.sohel@murdoch.edu.au}

\author{Puneet Goyal}
\orcid{0000-0002-6196-9347}
\affiliation{%
  \institution{Indian Institute of Technology Ropar}
  \city{Rupnagar}
  \state{Punjab}
  \country{India}
}
\email{puneet@iitrpr.ac.in}

% Running header for a long author list.
%\renewcommand{\shortauthors}{Dhar et al.}

\renewcommand{\shortauthors}{Joy Dhar et al.}


%\renewcommand{\shortauthors}{Dhar et al.}

\begin{abstract} 
Multimodal fusion learning (\texttt{MFL}), a framework for jointly learning from heterogeneous data sources, has shown great potential in fields such as medicine, science, and engineering.
It is extremely desirable in the medical domain, where we are faced with disparate data modalities such as imaging, clinical records, and omics. 
However, existing \texttt{MFL} strategies face several major challenges.  
First, they struggle to capture complex cross-modal interactions effectively, which in turn limits performance improvements. 
Second, they incur high computational costs, restricting their applicability in resource-constrained healthcare~\texttt{AI} applications.
Finally, they are often designed and evaluated for narrow, fixed modality configurations (e.g., imaging-only, or specific pairs such as image and omics), which limits evidence of their adaptability and generalizability to broader collections of heterogeneous medical modalities. 
To address these challenges, we propose a novel \texttt{MFL} framework --
\textbf{C}ascaded \textbf{U}nified \textbf{R}epresentation Learning for \textbf{E}fficient Fusion Network (\texttt{CURE}) -- a lightweight and scalable framework that progressively integrates various modalities through a novel efficient Hybrid Geometry Aware Fusion layer (\texttt{HyFuse}), where each \texttt{HyFuse} layer is sequentially learned for each modality, making the framework adaptable and generalizable. 
Within \texttt{HyFuse}, an efficient residual convolution module captures rich multi-scale features to ensure cost-effective learning, while a hybrid-space aware attention mixer learns coarse-to-fine structural cues to better preserve cross-modal relationships.   
Complementary learnable late-fusion and shared-information refinement modules are then employed to learn robust modality-order-invariant shared representations, which in turn yields consistent performance improvements. 
Extensive evaluations on $16$ public datasets show that \texttt{CURE} outperforms leading multimodal fusion methods (e.g., DRIFA-Net and HEALNet), boosting performance by up to $\approx3.97\%$ and lowering computational costs by up to $\approx87.8\%$, ensuring more effective and reliable predictions. 
\textcolor{blue}{Code:~\url{https://github.com/misti1203/CURE}.}
%\vspace{-0.5cm}
\end{abstract}



%\ccsdesc[500]{Do Not Use This Code~Generate the Correct Terms for Your Paper}
%\ccsdesc[300]{Do Not Use This Code~Generate the Correct Terms for Your Paper}
%\ccsdesc{Do Not Use This Code~Generate the Correct Terms for Your Paper}
%\ccsdesc[100]{Do Not Use This Code~Generate the Correct Terms for Your Paper}



\maketitle

%\vspace{-0.3cm}

%%%%%%%%%%%%%%%%%%%%%%%%%%%%%%%%%%%%%%%%%%%%%%%%%

%%%%%%%%%%%%%%%

%%%%%%%%%%%%%%%%
%%%%%%%%%%%%%%%%%%
\vspace{-0.1cm}
\section{Introduction} \label{intro}
%%%%%%%%%%%

\begin{figure}
\centering
  \includegraphics[width=0.48\textwidth]{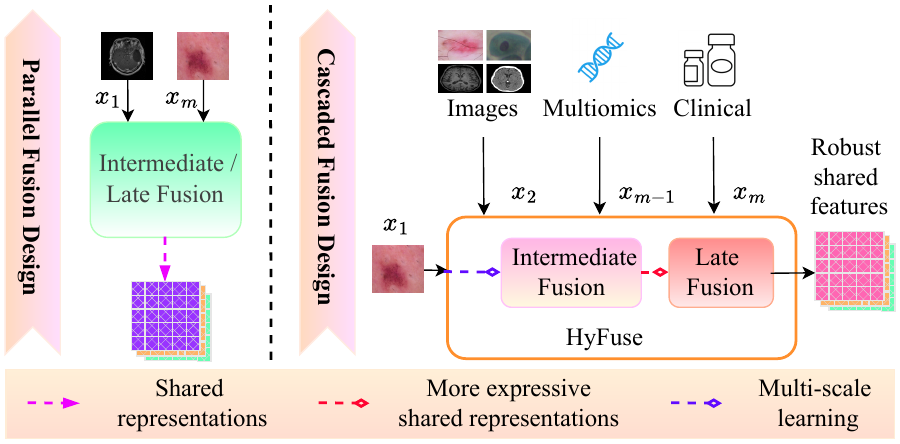}
      \vspace{-0.7cm}
  \caption{\small
Comparative high-level overview of various \texttt{MFL} methods -- 
\textbf{(Left)} architecture of intermediate or late fusion methods (e.g., \texttt{DRIFA-Net}~\cite{dhar2024multimodal}, 
\texttt{MuMu}~\cite{islam2022mumu},  
\texttt{MOTCAT}~\cite{xu2023multimodal})
%, leading to high computational costs and limited adaptability to heterogeneous modalities.  
%\textbf{(B)} Architecture of hybrid early and intermediate fusion methods (\texttt{HEALNet}
%~\cite{hemker2024healnet}
%).
and \textbf{(Right)} architecture of our proposed \texttt{CURE} framework.}  
  \label{fig:fig1}
  \vspace{-0.4cm}
\end{figure}

%%%%%%%%%%%

Multimodal fusion learning (\texttt{MFL}) remains a key challenge in machine learning \citep{islam2022mumu, islam2020hamlet, zhou2023cacfnet, dhar2024multimodal, hemker2024healnet, hemker2025multimodallegomodelmerging, dhar2026hypca, dhar2026effective}. It is of great importance in healthcare research, as there is a strong need to integrate heterogeneous medical data from disparate modalities---such as radiology, dermoscopy, multi-omics, and electronic health records (\texttt{EHR}s)---for diverse downstream tasks, including multi-disease analysis \citep{dhar2024multimodal, dhar2025towards, hemker2025multimodallegomodelmerging,dhar2026certified,  dhar2024uncertainty, hemker2024healnet}. 
In practice, these modalities represent data at different scales, are often unpaired, and can follow distinct, non-overlapping distributions \cite{hemker2025multimodallegomodelmerging}. Although modern multimodal fusion architectures \cite{lai2024carzero} can excel when modalities are \emph{paired} (e.g., audio--visual--text) \citep{goyal2016towards}---they typically rely on expensive end-to-end joint training and assume access to paired data, which is often scarce in biomedical applications \cite{hemker2025multimodallegomodelmerging}.   This motivates an open research problem in medical multimodal fusion paradigms: \emph{\textbf{how can we design a state-of-the-art multimodal fusion architecture that learns robust shared representations from heterogeneous medical modalities while remaining cost-effective?}}   

To address this question, classical multimodal fusion paradigms---early fusion\footnote{Early fusion concatenates raw modalities at the input, enabling end-to-end learning but often obscuring modality-specific cues, which undermines robustness~\citep{chen2020pathomic}.}~\citep{jaegle2021perceiver}, late fusion\footnote{Late fusion aggregates modality-specific outputs to preserve structural cues but limits cross-modal interaction while increasing computational cost~\citep{liang2022foundations}.}~\citep{islam2022mumu, islam2020hamlet, cheng2022fully, huang2021gloria}, and intermediate fusion\footnote{Intermediate fusion pipelines learn shared representations but can conflate modality-specific and modality-shared semantics; this strategy is, however, computationally expensive~\citep{cui2023deep}.}~\citep{dhar2024multimodal, xu2023multimodal} (\textcolor{black}{\textbf{see Fig.~\ref{fig:fig1}}})---have been widely adopted and remain the most dominant approaches for multimodal fusion. Despite their promise, broader adoption of multimodal fusion in \texttt{AI}-driven healthcare is still hindered by three main challenges raised by the above question---\textbf{performance}, \textbf{efficiency}, and \textbf{generalization}. 

\medskip
The \textbf{performance} challenge stems from the fact that existing architectures often lack an effective fusion mechanism. In particular, the lack of specialized attention modules or architectural inductive biases that encourage representational diversification restricts their capacity to preserve coarse-to-fine structural cues; this, in turn, constrains their ability to model complex cross-modal interactions. Consequently, they often \emph{learn shared representations with limited expressiveness, which hampers multimodal fusion performance improvements} {(\textbf{\emph{\textcolor{blue}{Challenge 1}}})\label{C1}}. 
The \textbf{efficiency} issue stems from the fact that many existing fusion methods rely on large, \emph{high-dimensional attention matrices or computationally intensive architectures}. This imposes \emph{significant computational overhead, limiting the applicability} of these methods in resource-constrained medical~\texttt{AI} applications (\textbf{\emph{\textcolor{blue}{Challenge 2}}}). 
The \textbf{generalizability} concern stems from the fact that existing multimodal fusion frameworks are often restricted in their ability to scale across diverse medical data sources, as they are mostly trained on a \emph{limited set of paired modalities rather than learning from heterogeneous medical data that include both paired and unpaired modalities} (\textbf{\emph{\textcolor{blue}{Challenge 3}}}). Moreover, existing works often lack the capability to leverage heterogeneous medical data within a single training cycle due to increased computational overhead. Ideally, one should adopt a strategy that leverages heterogeneous medical modality data within a single training cycle to reduce computational cost and ensure better generalization. So, in a way, the challenge of generalization is related to that of performance and efficiency.   

%As a result, learning robust shared representations across such heterogeneous modalities at low computational cost remains an open research problem. These observations motivate the following central question:  %\vspace{-0.5cm} %\begin{quote} %\emph{Can we design a state-of-the-art multimodal fusion framework that captures robust shared features while balancing the performance--efficiency trade-off?} %\end{quote} %\vspace{-0.4cm}     

%%%%%%%%%%%%%%   

\begin{figure}[t]     \centering     \includegraphics[width=0.48\textwidth]{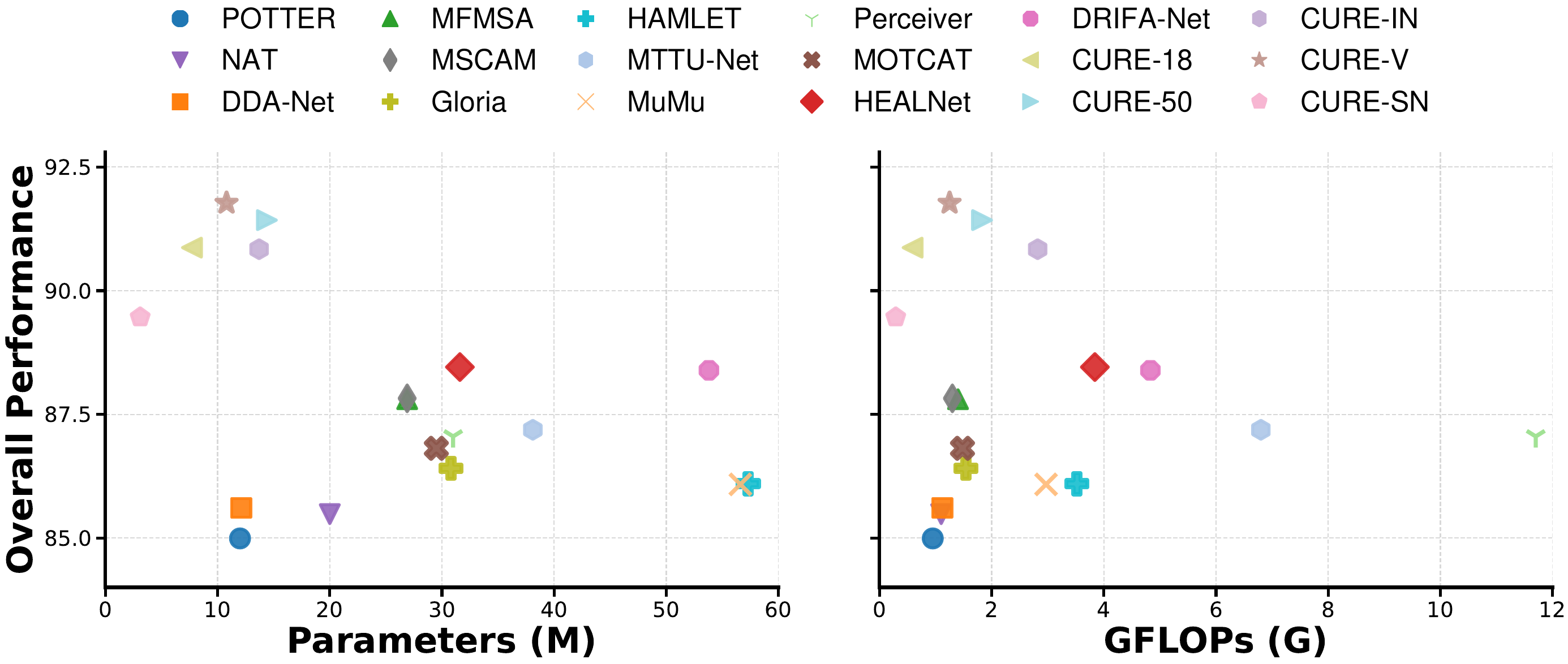}     
\vspace{-0.7cm}     
%\caption{\small Computational complexity comparison: overall performance (i.e., \texttt{ACC} + \texttt{AUC} + \texttt{C--Index}) vs. GFlops and parameters across heterogeneous unpaired modalities datasets (Tab. \ref{tab:tab1}).}     
\caption{\small Computational complexity comparison: overall performance (i.e., \texttt{ACC} + \texttt{AUC} + \texttt{C--Index}) vs. \texttt{GFLOPs}
and parameters across heterogeneous unpaired modality datasets (Tab.~\ref{tab:tab1}).}
%\caption{Mean percentage uplift of multimodal fusion models over the best unimodal baselines across heterogeneous medical datasets. \texttt{CoVANT} variants outperforms prior leading multimodal fusion baselines. }     \vspace{-0.5cm}     
\label{fig:fig2}     
%\vspace{-0.2cm} 

\end{figure}   %%%%%%%%%%%%%%   

\medskip
To address these challenges, we propose \textbf{C}ascaded \textbf{U}nified \textbf{R}epresentation Learning for \textbf{E}fficient Fusion Network (\texttt{CURE}) -- a novel framework of multimodal fusion learning to \textit{enable \textbf{effective}, \textbf{efficient}, \textbf{generalizable} and \textbf{scalable} integration in resource-constrained \texttt{AI}-driven healthcare applications}. The main component of~\texttt{CURE} is an \textbf{Efficient Hybrid Geometry Aware Fusion} (\texttt{HyFuse}) layer that exploits \emph{hybrid intermediate and late fusion} strategies -- effectively \emph{balancing optimal performance with minimal computational cost} (\textbf{see Fig. \ref{fig:fig2}}). This design facilitates the learning of modality order invariant robust shared representations and ensures scalability across any number of heterogeneous modalities for \emph{improved generalization}.   \texttt{CURE} employs a single-pass cascaded fusion pipeline, where modalities are processed sequentially and incrementally fused through successive \texttt{HyFuse} layers (\textbf{Fig.~\ref{fig:fig1}}).  Unlike prior methods \citep{dhar2024multimodal, hemker2025multimodallegomodelmerging, islam2020hamlet, islam2022mumu} that require additional encoders in parallel to incorporate each modality input---resulting in high-cost architectures (\textcolor{black}{\textbf{as shown in Figs.~\ref{fig:fig1}--\ref{fig:fig2}}})---our approach eliminates these computational bottlenecks.

\medskip
In our proposed framework, two modalities are first processed by a \texttt{HyFuse} layer, which captures \emph{multi-scale cues to encourage representational diversity}. These representations are further refined in a~\texttt{HyFuse} layer using a \textit{hybrid-space aware attention mixer mechanism}, that \emph{models coarse-to-fine cross-modal interactions while preserving the complex structural properties} of each modality. Next, a learnable late-fusion gating strategy is used that integrates the refined outputs into a shared representation -- thereby \emph{learning robust shared features}. This information (shared representation), along with the next input modality, is then fed into the subsequent \texttt{HyFuse} layer. In parallel, an intermediate-fusion pathway with a shared-information refinement module further aggregates hybrid-space shared features across modalities. This design \emph{promotes modality-order-invariant shared-feature learning and supports seamless integration of an arbitrary number of heterogeneous modalities}.   This process repeats until all input modalities are processed. The main contributions of our work are as follows: 

%%%%%%%%%%%%%%%%%%%%%%%%%%%%%%%%%%%%   

%%%%%%%%%%%%%%%%%%%%%%%
%\vspace{-0.3cm}
\begin{figure*}[t]
    \centering
    \includegraphics[width=\textwidth]{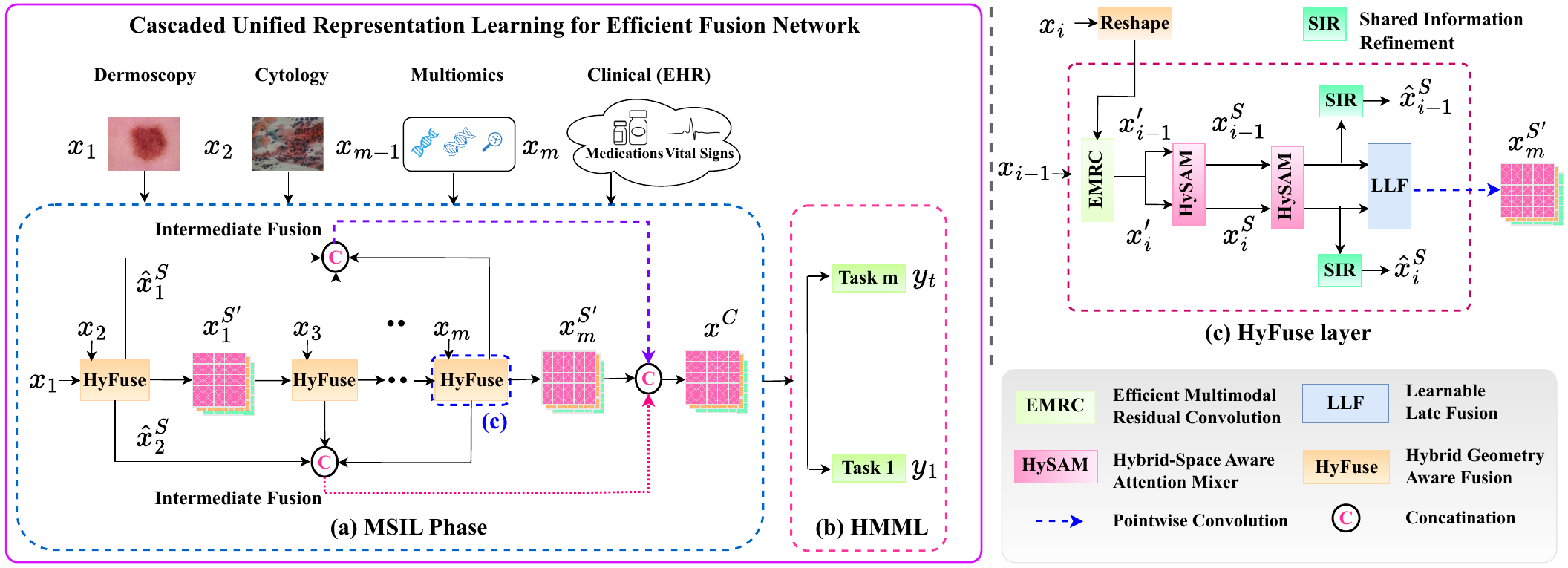}
    \vspace{-0.7cm}
     \caption{\small Overview of \texttt{CURE} framework, comprising the~\texttt{MSIL} and~\texttt{HMML} phases.  \textbf{(a)} Overview of \texttt{MSIL} phase utilizing \texttt{HyFuse} layer.
    \textbf{(b)} Overview of \texttt{HMML} phase.
    \textbf{(c)} Layout of~\texttt{HyFuse} layer, composed of \texttt{EMRC}, \texttt{HySAM}, \texttt{LLF}, and \texttt{SIR} modules, to learn robust shared features ($x^{S'}_j$); this, in turn, yields modality-order-invariant shared representations $x^{C}$ via the intermediate fusion in \textbf{(a); See Sec. \ref{LF_strategy}--\ref{shared_sir})}.}
    \label{fig:fig3}
    \vspace{-0.0cm}
\end{figure*}
%\vspace{-0.0cm}

%\vspace{-0.2cm}  
\medskip
\begin{itemize}[leftmargin=*, label=\textbullet, itemsep=0pt, topsep=0pt]  

\item We introduce \textbf{\texttt{CURE}}, a novel light-weight end‑to‑end multimodal fusion framework that addresses the challenges of \textbf{\emph{performance}}, \textbf{\emph{efficiency}}, and \textbf{\emph{generalizability}} in existing frameworks --  making it well-suited for \texttt{AI}-driven healthcare applications with many modalities in low-resource environments.     

\item We propose a \textbf{novel~\texttt{HyFuse} layer} that is based on a hybrid-space aware attention mixer mechanism, utilizing an intermediate fusion strategy by integrating \emph{\textbf{hyperbolic}} and \emph{\textbf{quantum‑inspired}} embeddings. As a result, it learns much richer spatial and frequency‑domain representations, preserves \emph{coarse-to-fine modality‑specific structural details}, and captures \emph{complementary cross‑modal dependencies}.    

\item We conduct extensive evaluations against state-of-the-art methods across \textit{\textbf{sixteen heterogeneous medical datasets}}, demonstrating significant performance improvements.  
\vspace{0.05cm} 

\end{itemize}
%%%%%%%%%%%

%\vspace{-0.3cm}
\section{Related Work}
%\vspace{-0.3cm}

We focus on multimodal learning problems in biomedical data, where the modalities are structurally heterogeneous -- i.e., using imaging modalities (e.g., dermoscopy, Pap smear) along with tabular data (e.g., multi-omics and clinical records). 
This is in contrast with existing works that are limited to homogeneous modalities only.
%(e.g., imaging only or omics only fusion). 
While multimodal fusion learning strategies based on early, late, intermediate, and hybrid methods have driven significant research for homogeneous modalities (e.g., natural vision tasks~\citep{Abdelhalim2021,peng2022balanced,wang2020deep,wang2020learning,joze2020mmtm,ma2021smil})
%in multimodal fusion learning models, particularly in natural vision tasks 
, their adoption for structurally heterogeneous modalities (e.g., biomedical data) remains limited
~\citep{cheng2022fully}. 
As we discussed earlier, each strategy of multimodal learning has its own strengths and weaknesses. E.g., early fusion methods, such as \texttt{Perceiver}~\citep{jaegle2021perceiver}, naïvely concatenate raw inputs and apply iterative self and cross-attention layers for end-to-end learning. However, such naïve fusion dilutes the modality-specific signals and increases feature dimensionality
~\citep{wang2020learning}. 
Late fusion methods (e.g., \texttt{MTTU-Net}~\citep{cheng2022fully}, \texttt{GLORIA}~\citep{huang2021gloria}, etc.) encode each modality independently -- combining \texttt{CNN}s and transformers for glioma segmentation or aligning images with radiology reports via global local attention.
Its variants like \texttt{HAMLET}~\citep{islam2020hamlet} and \texttt{MuMu}~\citep{islam2022mumu} include multi-head self-attention for richer context learning. Despite preserving modality-specific structure, they fail to capture fine-grained cross-modal interactions, incur progressive information loss from cascaded attention layers, and remain computationally intensive due to heavy convolutional and attention operations.
Intermediate fusion methods (e.g., \texttt{CAF}~\citep{He2023}, \texttt{DRIFA-Net}~\citep{dhar2024multimodal},  \texttt{MOTCAT}~\citep{xu2023multimodal}, etc.), employ cross-modal, cascaded dual attention (or optimal transport co-attention) to fuse modalities within their attention module. 
While more expressive, they often impose modality-specific training constraints and incur substantial computational cost, which can limit robustness and scalability.
Hybrid designs such as~\texttt{HEALNet}~\citep{hemker2024healnet} bridge early and intermediate fusion through a hybrid early fusion layer to enrich shared representations. More recent architectures, including \texttt{CA-MLIF}~\citep{an2025mlif} and \texttt{MMLego}~\citep{hemker2025multimodallegomodelmerging}, respectively interleave cross-attention with low-rank interaction fusion and compose pre-trained unimodal encoders with lightweight fusion blocks to better model cross-modal dependencies. Nevertheless, many existing approaches still rely on high-dimensional attention matrices and bespoke fusion mechanisms, leading to high computational cost and limiting practicality in low-resource healthcare \texttt{AI} settings. \emph{\textbf{App.~\ref{related} details the gap analysis and our rationale for hybrid spaces in CURE.}}

%%%%%%%%%%%%

%Several recent methods for clinical prognosis explicitly model the interplay between modality-specific and modality-shared representations. For example, Pathomic Fusion \citep{chen2020pathomic} employs separate encoders for histology and genomics and fuses them via Kronecker products and gated attention, yielding distinct unimodal and pairwise interaction terms for cancer diagnosis and survival prediction. \citep{steyaert2023multimodal} and \citep{cui2023deep} systematically evaluate early, intermediate, late and hybrid fusion strategies on \texttt{TCGA}-style cohorts, emphasizing that preserving modality-specific structure while learning a shared latent space is crucial for biomarker discovery. 

%%%%%%%%%
\medskip
\noindent\textbf{Our Approach:} \texttt{CURE} overcomes these limitations with the following key innovations. First, it has a modular design that \textit{scales seamlessly to any number of modalities} by adding a novel \texttt{HyFuse} layer, which captures complex \textit{cross-modal interactions} while preserving each \textit{modality’s structural details}, thereby learning \textit{robust shared representations}. Secondly, it has a novel module that leverages hyperbolic and quantum attentions in parallel and fuses their outputs to prevent \textit{progressive information loss} and capture \textit{complex hierarchical relationships} across modalities. 
\texttt{CURE} delivers improved \textit{generalizability} across diverse medical modalities and more holistic explanations than existing state-of-the-art (\texttt{SOTA}) fusion methods. 

%%%%%%%%%%%%%%%%%%%%%%%

\begin{figure*}[t]
    \centering
    \includegraphics[width=0.8\textwidth]{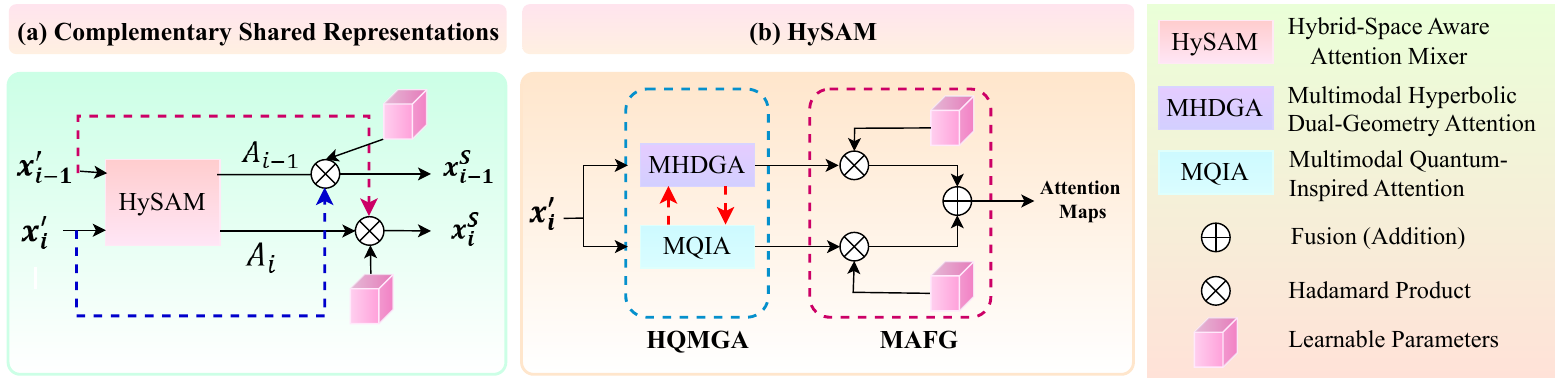}
    \vspace{-0.3cm}
    %\caption{Architecture of \texttt{EDF-Net}, consisting of two stages: (A) \texttt{MIFL}, which captures enhanced complementary shared representations; and (B) \texttt{MMTL}, responsible for multi-disease classification. \texttt{MIFL} incorporates three core blocks: (C) \texttt{SRCL}, for learning refined modality-specific representations--\(x^{\prime}_{i \in [1, m]} \); \texttt{ESFIA}, for jointly optimizing spatial and frequency information; and \texttt{CBIR}, for enhancing shared representational diversity. Within \texttt{SRCL}, (D) the \texttt{MSGCA} module includes (E) \texttt{MSHC}, with channel shuffle, and (F) \texttt{CIFA} for multi-scale feature refinement.}
    \caption{\small
(a) Illustration of the \texttt{HySAM} module’s role in learning more expressive shared representations $x^S_i$ for each $i \in \{1, \dots, m\}$. 
(b) Overview of the \texttt{HySAM} module, which employs the \texttt{HQMGA} block -- comprising the \texttt{MHDGA} and \texttt{MQIA} mechanisms and a \texttt{MAFG} block that fuses the attention maps from \texttt{MHDGA} and \texttt{MQIA} to learn hybrid-space aware shared attention maps $A_i$.
}
    \label{fig:fig4}
    \vspace{-0.2cm}
\end{figure*}

%%%%%%%%%%%%%%%%%%%%%%%%%%%%%%%%%%%%%%%%%%%%%%

%\vspace{-0.25cm}
\section{Proposed Method} \label{proposed}
%\vspace{-0.2cm}
\noindent\textbf{Problem Formulation: } %In this section, let us discuss our proposed \textbf{E}fficient \textbf{H}ybrid-fusion \textbf{P}hysics-informed \textbf{A}ttention \textbf{L}earning \textbf{N}etwork, called \texttt{EHPAL-Net}, that 
\texttt{CURE} enhances \texttt{MFL} through cascaded spatial- and frequency-domain information integration. Let \( \mathcal{X} = \{x_i\}_{i=1}^m \) denote \(m\) heterogeneous input modalities, where each \(x_i \in \mathbb{R}^{H_i \times W_i \times C_i} \cup \mathbb{R}^{D_i}\) represents either imaging data (with spatial dimensions height \(H_i\), width \(W_i\), channels \(C_i\)) or non-imaging data (e.g., multi-omics or \texttt{EHR}) as \(D_i\)-dimensional vectors. 
Let \( \mathcal{Y} = \{y_j\}_{j=1}^t \) denote \(t\) multi-task labels. 
Notably, non-imaging vectors are reshaped into pseudo-image tensors in $\mathbb{R}^{H_i \times W_i \times C_i}$ to enable uniform fusion.
Our aim is to learn a function -- \( \mathcal{F}(\cdot) \) that captures modality order invariant robust shared representations: $x^C$ %\( X^S = \{x^{s}_{i}\}_{i=1}^{m} \) 
through the mapping \( \mathcal{F}:\mathcal{X}\;\longrightarrow\;\mathcal{Y} \), optimizing both performance as well as computational efficiency.

%\textcolor{purple!60}{In this work, each $x_i$ corresponds to one dataset‑level modality stream. For example, we use dermoscopy, cytology, histology tiles, multi‑omics feature vectors, and clinical time‑series. Although the notation \( \mathcal{X} = \{x_i\}_{i=1}^m \) suggests a common index set, in practice these modalities are unpaired: the underlying subjects and label spaces are disjoint across datasets. The \texttt{MSIL} phase therefore learns a shared representation $x^C$ across heterogeneous, unpaired modalities, while the \texttt{HMML} phase attaches task‑specific heads to each modality.}
 
%\vspace{-0.1cm}

\medskip
\noindent\textbf{Method Overview: } 
%We present a holistic overview of \texttt{EHPAL-Net} framework (ref. Fig.~\ref{fig:fig2}), which 
\texttt{CURE} differs from conventional \texttt{MFL} methods 
%that rely on a single fusion function (e.g., early \cite{jaegle2021perceiver}, late \cite{islam2022mumu, islam2020hamlet, joze2020mmtm, wang2020learning}, or intermediate \cite{joy_wacv, xu2023multimodal}) for two modalities during training. In contrast, our \texttt{EHPAL-Net} 
as it leverages a sequential multimodal integration pipeline, where modality-specific inputs are processed sequentially and incrementally fused through its novel \textbf{Efficient Hybrid Geometry Aware Fusion} (\texttt{HyFuse}) layer. 
%By integrating dual-geometry and quantum information with convolutional neural networks, it captures spatial and frequency information across various modalities. %\st{, enabling robust complementary shared representation learning and optimizing performance-computation tradeoffs for heterogeneous medical data.} 
\texttt{CURE} comprises two salient phases: 
(1) \textbf{Multimodal Shared Information Learning} (\texttt{MSIL}) -- a sequential integration of heterogeneous modalities to learn modality order invariant robust shared features, and
(2) \textbf{Heterogeneous Modality-Specific Multitask Learning} (\texttt{HMML}) -- utilizes the shared representations for multiple tasks, such as multi-disease classification and patient survival prediction.
In the following, we will delve into the details of these two phases.
%\emph{\textcolor{purple}{An overview of our proposed method is shown in Figure~\ref{fig:fig2}}}. \emph{\textcolor{purple}{Detailed algorithm of~\texttt{CURE} is given in Algorithms \ref{al:al1}-\ref{al:al2} in Appendix~\ref{algorithm}}}.
\textcolor{black}{\textbf{\emph{An overview of the implementation of \texttt{CURE} is given in Fig.~\ref{fig:fig3}, and pseudo-code is provided in Algorithms~\ref{al:al1}-\ref{al:al2}}}}.

%\texttt{EHPAL-Net} employs an iterative learning framework to process heterogeneous modalities. 
 
%\vspace{-0.2cm} 
\subsection{Multimodal Shared Information Learning} \label{sec_MSIL} 
\texttt{MSIL} phase utilizes \texttt{HyFuse} layers to learn modality-order-invariant robust shared features over $m$ modalities. 
At \texttt{HyFuse} layer $i$, the current and previous modality inputs ($x_{i}$ and $x_{i-1}$)
%; e.g., $i=1,2$ in the first layer, then $i+1$ thereafter) 
are processed together through four core modules, namely a) \textbf{Efficient Multimodal Residual Convolution (\texttt{EMRC})}, b) \textbf{Hybrid-Space Aware Attention Mixer (\texttt{HySAM})}, c) \textbf{Learnable Late Fusion (\texttt{LLF})}, and d) \textbf{Shared Information Refinement (\texttt{SIR})} modules.
We will delve into the details of these four modules in the following. 
It is important to note that each \texttt{HyFuse} layer uses two \texttt{HySAM} modules to refine shared representations, then fuses their outputs through an \texttt{LLF} strategy, i.e., leveraging trainable weights to capture more expressive multimodal information \textcolor{black}{\textbf{(Fig.~\ref{fig:fig3})}}. 
This representation, along with the next modality’s input, is fed to the subsequent \texttt{HyFuse} layer. It can be seen that \texttt{MSIL} phase is a single-pass sequential (cascaded) pipeline that processes modalities in an order-invariant manner.
%\st{This enable efficient cross-modal knowledge transfer while preserving hierarchical, modality-specific structural details.} 
This design facilitates modality-order-invariant robust shared features learning by effectively optimizing performance and computational cost. \textcolor{black}{\textbf{\emph{A detailed view of the HyFuse implementation is shown in Fig.~\ref{fig:fig3} (c)}}}.

%%%%%%%%%%%%%%%%%%%%%%%%%%%%%%%%%%%%%%%

%\vspace{-0.2cm}
\subsubsection{\textbf{Efficient Multimodal Residual Convolution Module}} \label{EMRC}

%We design the Efficient Multimodal Residual Convolution (\texttt{EMRC}) module (ref. \ref{Preliminaries}) to capture multi-scale spatial representations while ensuring low computational cost. The \texttt{EMRC} incorporates Modality-specific Heterogeneous Convolutions Fusion (MHCF) blocks, along with BatchNorm ($\texttt{B}(\cdot)$), ReLU ($\mathcal{R}(\cdot)$), and pointwise convolution ($\texttt{PC}(\cdot)$) to facilitate progressive refinement. Unlike the uniform depth-wise convolutions employed in \texttt{EMCAD}'s \texttt{MSDC} \cite{rahman2024emcad}, \texttt{MHCF} uses diverse branch-wise convolutions—Group-Pointwise (\texttt{GPC}), Dilated Depth-wise (\texttt{DDC}), and Depth-wise (\texttt{DWC})—at varying scales (\(1\!\times\!1\), \(3\!\times\!3\), \(5\!\times\!5\)), capturing heterogeneous spatial contexts and promoting branch-wise heterogeneity. The resulting contexts are fused and refined using a channel shuffle ($\texttt{C}(\cdot)$) to facilitate inter-channel communication, thereby enhancing representational diversity, denoted as $x_{i}^{\prime} \in \texttt{EMRC}(\cdot)$: 
%The \texttt{EMRC} is formally defined:

We design the Efficient Multimodal Residual Convolution (\texttt{EMRC}) module that captures multi-scale spatial representations \(x_i^{\prime} \) for each modality while ensuring cost-effective computation. 
To achieve this, \texttt{EMRC} incorporates Modality-specific Heterogeneous Convolutions Fusion (\texttt{MHCF}) blocks, which facilitate progressive refinement of modality-specific inputs. 
\textcolor{black}{\textbf{\emph{The architectural and implementation details of the \texttt{EMRC} module are provided in App.~\ref{emrc}}}}.

%%%%%%%%%%%%%%%%%%%%%%%
%\vspace{-0.2cm}

%\vspace{-0.2cm}

%%%%%%%%%%%%%%%%%%%%%%%%%%

%%%%%%%%%%%%%%%%%%%%%%%
%\vspace{-0.2cm}
\begin{figure*}[t]
    \centering
    \includegraphics[width=0.9\textwidth]{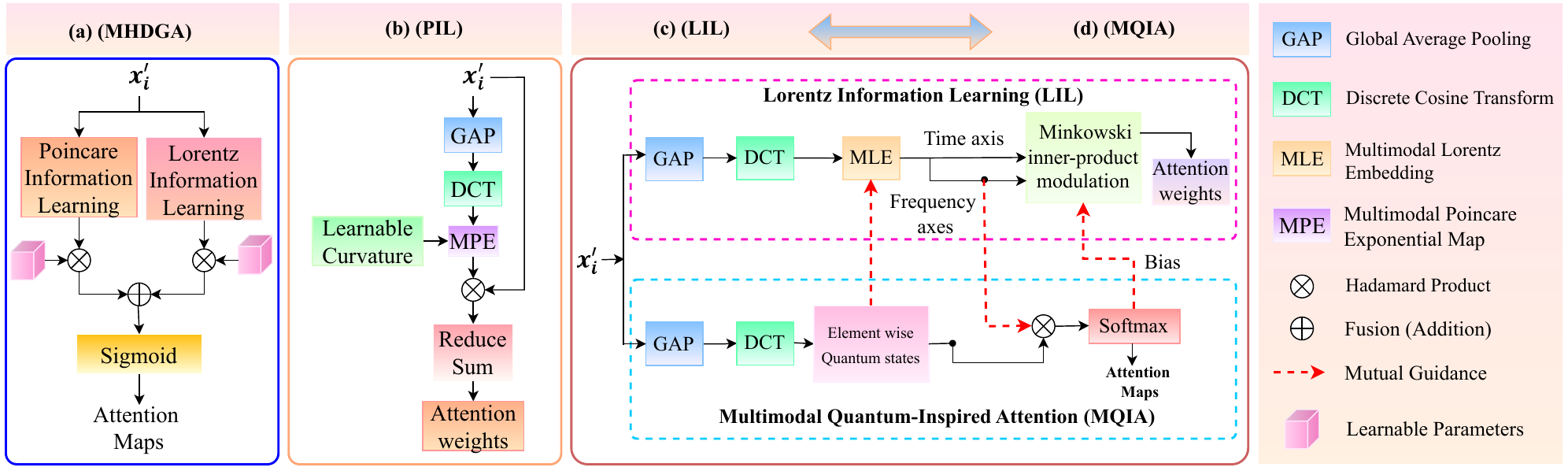}
    \vspace{-0.3cm}
      \caption{\small
(a) Overview of the \texttt{MHDGA} module, which incorporates \texttt{PIL} and \texttt{LIL} to learn \emph{dual-geometry-aware} attention maps.   
(b) Detailed view of the \texttt{PIL} branch for computing Poincar\'e attention weights.  
(c--d) Overview of the \texttt{LIL}--\texttt{MQIA} \emph{mutual-guidance} loop (red arrows), which couples Lorentzian hyperbolic embeddings with quantum-inspired interactions to yield richer, co-evolving attention weights.  
Concretely, \texttt{LIL} embeds features in the Lorentz manifold via \texttt{MLE} (Eqs.~\ref{eq:eq6}--\ref{eq:eq7}), while \texttt{MQIA} projects quantum states through the same embedding to produce a channel-wise bias (Eq.~\ref{eq:eq9}) that feeds back to refine Lorentzian attention (Eq.~\ref{eq:eq8}), yielding co-adapted attention weights.}
    \label{fig:fig5}
    \vspace{-0.3cm}
\end{figure*}

%\vspace{-0.2cm}

%%%%%%%%%%%%%%%%%%%%%%%%%%

\begin{algorithm}[H]
\caption{\texttt{CURE($x_i$)}}
\label{al:al1}
%\begin{algorithmic}[1]
\begin{algorithmic}[1]
\STATE \textbf{Input:} Modalities, $x_i = [x_1, x_2, \ldots, x_m]$, where $x_i \in \mathbb{R}^{H_i \times W_i \times C_i} \cup \mathbb{R}^{D_i}$
\STATE \textbf{Output:} Shared representation $x^C$ and downstream multitask outputs
\STATE \textbf{Procedure:}
\IF{phase == \texttt{MSIL}}
    \STATE \textit{/* MSIL phase for learning robust representations */}
    \FOR{each $x_i$, where $i = 1$ to $m$}
        \IF{$i == 2$}
            \STATE $(\hat{x}^S_i,\,x^{S^\prime}_i) \gets$  \texttt{HyFuse}($x_{i-1},\,x_i$)
        \ELSE
            \STATE $(\hat{x}^S_i,\,x^{S^\prime}_i) \gets$  \texttt{HyFuse}($x^{S'}_i,\,x_{i+1}$)
        \ENDIF
        \IF{$i == m$} 
            \STATE \textit{/* Modality order invariant robust shared features Eq.~\ref{eq:eq10} */}
            \STATE $x^C \gets \theta(x^{S'}_i,\, \theta(\hat{x}^S_i))$ 
        \ENDIF
    \ENDFOR
\ELSE
    \STATE \textit{/* HMML phase for multitask learning */}
    \FOR{each task $t$}
        \STATE $\mathcal{L}_{\texttt{HMML}} \gets \sum_{t=1}^{T} \sum_{m=1}^{M} \lambda_t^M \cdot \mathcal{L}_t^m(\mathcal{F}(x^C;\beta),\, \mathcal{Y})$
        \STATE $\beta^* \gets \arg \min_\beta \left( \mathcal{L}_{\texttt{HMML}} \right)$        \quad  // as per Eq. \ref{eq:eq11} 
    \ENDFOR
\ENDIF
\end{algorithmic}
\end{algorithm}

%%%%%%%%%%%%%%%%%%%%%%%%%%%%%%%%%%%%%%
%\vspace{-0.2cm}
\subsubsection{\textbf{Hybrid-Space Aware Attention Mixer Module}} \label{pcmfa} 
%\vspace{-0.2cm}

Recent advances in hyperbolic neural networks (\texttt{HNN}s) and quantum neural networks (\texttt{QNN}s) have demonstrated their ability to learn \emph{richer representations than traditional Euclidean architectures} (e.g., \texttt{CNN}s, Transformers), especially on complex natural visual tasks \citep{ganea2018hyperbolic, peng2021hyperbolic, Cong2018QuantumCNN, shi2024qsan}. \texttt{HNN}s leverage non-Euclidean geometries such as the Poincaré ball \citep{NickelKiela2017Poincare} and the Lorentz hyperboloid \citep{Chen2021FullyHyperbolicNeuralNetworks}, to \emph{preserve hierarchical structural information} efficiently, often in \emph{lower-dimensional spaces} than their Euclidean counterparts. Concurrently, \texttt{QNN}s employ quantum principles like superposition and entanglement to operate in high-dimensional Hilbert spaces \citep{Cong2018QuantumCNN}, with recent variants (e.g., \texttt{QSAN} \citep{shi2024qsan} and \texttt{QSANN} \citep{Li2022QSANN}) redefining neural layers (e.g., convolutions, attentions) using parameterized quantum states, \emph{improving representational capacity while potentially reducing cost}.

\textit{Despite their respective complementary strengths -- hierarchical structure preservation in \texttt{HNN}s and expressive representational capacity in \texttt{QNN}s -- these paradigms have not been unified effectively within a single framework to benefit \texttt{MFL}}. 
To bridge this gap, we propose a unified attention mechanism -- Hybrid-Space Aware Attention Mixer (\texttt{HySAM}) module, which is the \emph{key component} of our proposed~\texttt{CURE} framework.
Like \texttt{EMRC}, \texttt{HySAM} module (\textcolor{black}{\textbf{ref. Fig.~\ref{fig:fig4}}}) is designed to learn \emph{more expressive shared representations} (\( x^{s}_{i} \)), however, it does so by \emph{jointly optimizing cross-modal interactions} in hyperbolic and quantum spaces. 
This way, it captures \emph{rich structural relationships} by leveraging both non-Euclidean geometry and quantum states.
It processes the refined multimodal inputs $\{x^{\prime}_i\}_{i=1}^m$ (output of the \texttt{EMRC} module) and learns hybrid-space aware shared attention maps ( $A_{i}$). 
\texttt{HySAM} module consists of two core blocks: 
\medskip
\begin{itemize}[leftmargin=*, label=\textbullet, itemsep=0pt, topsep=0pt]  

\item \textbf{Hyperbolic Quantum Mutual Guidance Attention} (\texttt{HQMGA}) block, which captures complex hierarchical structures across modalities by employing mutually guided attention streams in hyperbolic and quantum spaces (\textcolor{black}{\textbf{Fig.~\ref{fig:fig4} (b) and }\textcolor{black}{\textbf{Fig. \ref{fig:fig5}(c--d)}}}), and \item  \textbf{Multimodal Attention Fusion Gating} (\texttt{MAFG}) block that fuses information from hyperbolic-quantum streams across modalities, facilitating the learning of refined shared representations.

\end{itemize}
\medskip
The working of these two blocks can be written in the following form:
%\vspace{-0.2cm} 
%\begin{itemize}[leftmargin=*] 
%\item Hyperbolic Quantum Mutual Guidance Attention \texttt{(HQMGA)} block: Captures complex hierarchical structures across modalities by employing mutually guided attention streams in hyperbolic and quantum spaces. 
%\vspace{-0.1cm} 
%\item Multimodal Attention Fusion Gating (\texttt{MAFG}) block: Fuses information from hyperbolic-quantum streams across modalities, facilitating the learning of refined complementary shared representations. 
%\end{itemize} 

\vspace{-0.25cm} 
\begin{equation}\small
A_i = \texttt{HySAM}(x^{s}_i, x^{\prime}_{i+1}) = \texttt{MAFG}\left( \texttt{HQMGA}\left(x^{s}_i, x^{\prime}_{i+1} \right) \right).   
\label{eq:eq1}
\end{equation}
where: 
%\vspace{-0.05cm}
\begin{equation}\small
%\quad x_i^s= \underbrace{x^{\prime}_{m+1-i} \odot \texttt{PCMFA}(x^{\prime}_i) \odot o_i}_{\textrm{Cross-modal Interactions}}  
%\quad x_i^{s} = x^{\prime}_{i-1} \odot A_{i-1} \odot \alpha_i}
\quad x_i^{S} = x^{\prime}_{i} \odot A_{i-1} \odot \alpha_i.
\label{eq:eq1_1}
\end{equation}
%where $x_i^s$ is the output of application of \texttt{PCMFA} on 
The above equation captures cross-modal interactions, where $\alpha_i$ represents modality-specific learnable parameters and $\odot$ denotes the Hadamard product.
%\textcolor{blue}{Joy, this equation is not clear to reader. what is m, what is i, give some context on indices. what is the output of PCMFA?}
Let us discuss \texttt{HQMGA} and~\texttt{MAFG} blocks in the following:
%\begin{itemize}[leftmargin=*]
%\item  

\medskip
\noindent\textbf{A. Hyperbolic Quantum Mutual Guidance Attention.} The \texttt{HQMGA} block is designed by integrating two main attention mechanisms \textcolor{black}{\textbf{(Fig. \ref{fig:fig4} (B))}} that we described in the following:

\medskip
\noindent \textbf{(1) Multimodal Hyperbolic Dual-Geometry Attention (\texttt{MHDGA})} mechanism \textcolor{black}{\textbf{(Fig.~\ref{fig:fig5} (a))}} that exploits complementary properties of the Poincaré ball and Lorentz models to learn rich hierarchical representations across modalities.
\texttt{MHDGA} mechanism is implemented with two sub-blocks -- \textbf{(i)} \emph{\textbf{Poincar\'e Information Learning}} (\texttt{PIL}($\cdot$)) and \textbf{(ii)} \emph{\textbf{Lorentz Information Learning}} (\texttt{LIL}($\cdot$)).
These two sub-blocks respectively compute Poincaré attention weights ($A^{P}_{i}$) and Lorentzian attention weights ($A^{L}_{i}$) \footnote{These components aim to preserve complex hierarchical structural cues across modalities in non-Euclidean spaces.}. 
The resulting geometry-specific weights are fused via learnable parameters $L$ and $P$, followed by a sigmoid activation ($\sigma(\cdot)$) to learn dual-geometry attention maps (for $x_i^{\prime}$ and $x_i^{s^\prime}$) -- in the following we will denote both $x_i^{\prime}$ and $x_i^{s^\prime}$ as $x_i^{\prime}$: 
%($A^{D}_i$), where $A^{D}_i = \texttt{MHDGA}(\cdot)$:
%\vspace{-0.5cm} 
\small{
\begin{eqnarray} %\small
A^{D}_i & = & \texttt{MHDGA}(x^{\prime}_i), \nonumber \\
& = & \sigma\bigl(L \times A_i^L + P \times A_i^P \bigr), \nonumber \\
& = & \sigma\bigl(L \times \texttt{LIL}(\psi_i) + P \times \texttt{PIL}(\psi_i) \bigr),
%\vspace{-0.1cm} 
\label{eq:eq3}
\end{eqnarray}}
Here $\psi_i \;=\;\texttt{GAP}\bigl(\texttt{DCT}(x^{\prime}_i)\bigr)\;\in\;\mathbb{R}^e$ represents the application of Global Average Pooling (\texttt{GAP}) followed by a Discrete Cosine Transform (\texttt{DCT}) to capture frequency components.
%: $\psi_i \;=\;\texttt{GAP}\bigl(\texttt{DCT}(x^{\prime}_i)\bigr)\;\in\;\mathbb{R}^e$.
%\vspace{-0.4cm}

\medskip
\noindent\textbf{(i) Poincaré Information Learning} sub-block \textcolor{black}{\textbf{(Fig.~\ref{fig:fig5} (b))}} is designed based on the principles of the Poincaré ball model~\citep{NickelKiela2017Poincare}.
%, which facilitates the preservation of hierarchical structural information in non-Euclidean space  
 Given multimodal frequency-domain inputs, i.e., $\psi_i$, we define~$\texttt{MPE}^{\tilde{c}}(\cdot)$ as Multimodal Poincar\'e exponential map with learnable curvature \(\tilde c\) and each \(\psi_i\) is projected onto the \(e\)-dimensional hyperbolic manifold with learnable curvature \(\tilde{c}\) as follows:
\begin{equation}  \small
\texttt{MPE}^{\tilde{c}}(\psi_i) = \tanh\!\left( \sqrt{\tilde{c}}\, \|\psi_i\| \right)\, 
\frac{\psi_i}{\|\psi_i\| + \varepsilon}.
\label{eq:eq4}
\end{equation}
%\vspace{-0.3cm} 
Here \(\|\cdot\|\) denotes the Euclidean norm and \(\varepsilon=10^{-6}\) ensures numerical stability. This ensures \(\|\texttt{MPE}^{\tilde c}(\psi_i )\|<1\), exploiting hyperbolic volume growth to compactly encode hierarchical relationships in low-dimensional manifolds.
To enable adaptive curvature \(\tilde c\), we modulate a base curvature \(c\) with fractal‐scaling weights \(\mathbf f\!\in\!\mathbb R^e\): 
$\tilde c \;=\; c \;\times\;\frac{1}{e}\sum_{j=1}^{e}\sigma\bigl(f_j\bigr)$. The base curvature is parameterized as: $c = \operatorname{clip}\bigl(e^k,\,0.1,\,10.0\bigr)$, with \(k\in\mathbb R\). This bounded formulation ensures that the geometry remains stable -- avoiding degenerate flatness (\(c\to0\)) or excessive sharpness (\(c\to\infty\)) -- across both Poincaré and Lorentz models.
 %Each \(\psi_i\) is projected onto the \(e\)-dimensional hyperbolic manifold with learnable curvature \(\tilde{c}\), and
 The Poincaré attention weights are computed via the Hadamard product between the hyperbolic projection and the original frequency components:

\vspace{-0.3cm} 
\begin{equation} \small
A_i^{P} = \texttt{PIL}(\psi_i) = \sum_{i} \left( \texttt{MPE}^{\tilde{c}}(\psi_i) \right) \, \odot \psi_i.
\label{eq:eq5}
\vspace{-0.1cm} 
\end{equation}

%We incorporate a learnable curvature \(\tilde c\) by modulating the base curvature \(c\) with fractal‐scaling weights \(\mathbf f\!\in\!\mathbb R^e\): $\tilde c \;=\; c \;\times\;\frac{1}{e}\sum_{j=1}^{e}\sigma\bigl(f_j\bigr)$. This \(\tilde c\) then replaces the fixed curvature in all hyperbolic mappings. To ensure stability, we parametrize the base curvature as: $c = \operatorname{clip}\bigl(e^k,\,0.1,\,10.0\bigr)$, with \(k\in\mathbb R\) a learnable log‐curvature. Clamping \(c\) to \([0.1,\,10.0]\) prevents the geometry from becoming too flat (\(c\to0\)) or too sharp (\(c\to\infty\)) in both Poincaré and Lorentz models.

%%%%%%%%%%%%%%%%%%%%%%%%%%%%%%%%%%%%%%%%%%%%%%%%
%\vspace{-0.4cm}

\medskip
\noindent\textbf{(ii) Lorentz Information Learning} sub-block \textcolor{black}{\textbf{(Fig.~\ref{fig:fig5} (c))}} 
%captures more effective hierarchical structural contexts in non-Euclidean spaces by learning richer frequency-temporal relationships in hyperbolic geometry.
is designed based on the Lorentzian hyperboloid model \citep{ganea2018hyperbolic}. 
Given multimodal frequency-domain inputs, i.e., $\psi_i$, we define Multimodal Lorentz Embedding denoted as $\texttt{MLE}(\psi_i)$, and split it into a temporal axis ($t_i$) and frequency components $(f_i)$ as follows:
%\vspace{-0.5cm} 
\begin{equation} \small
t_i,\, f_i = \underbrace{\texttt{MLE}(\psi_i)[:1]}_{\text{time axis}},\ \underbrace{\texttt{MLE}(\psi_i)[1:]}_{\text{frequency axes}}; \ \forall i \in [1, m].
\label{eq:eq6}
\end{equation}
Here we define $\texttt{MLE}(\psi_i)$ as:
\begin{equation} \small
    \texttt{MLE}(\psi_i) = \theta\left( \left( \frac{\sqrt{1 + \tilde{c} \, \|\psi_i\|_2^2}}{\tilde{c}} \right),\, \psi_i \right),
    \label{eq:eq7}
\end{equation}
%\vspace{-0.3cm} 
where $\theta$ represents concatenation. 
Next, we obtain a channel-wise bias $ (\delta_i \in\mathbb{R}^e) $ from the Multimodal Quantum-Inspired Attention (\texttt{MQIA}) mechanism (discussed next). 
This bias is injected into the Lorentz manifold through a Minkowski inner-product modulation \textcolor{black}{\textbf{(Fig.~\ref{fig:fig5} (c--d))}}, guiding the computation of Lorentzian attention weights:
%($A^{L}_{i} = \texttt{LIL}(\cdot)$): 

%\vspace{-0.6cm} 
\begin{equation} \small
A^{L}_i = \texttt{LIL}(t_i,f_i) =\;\alpha_i\!\Bigl(-\,t_i^2 + \sum_{l=1}^{i}\beta_l\,(f_{i,l} + \delta_{i,l})^2\Bigr), 
%\vspace{-0.1cm} 
\label{eq:eq8}
%\vspace{-0.1cm}
\end{equation}

where $\alpha_i$ and $\beta_l$ are learnable scalars modulating the contributions of temporal and frequency components. 

%\item 

%%%%%%%%

%%%%%%%

\medskip
\noindent \textbf{(2) Multimodal Quantum-Inspired Attention (\texttt{MQIA})} mechanism \textcolor{black}{\textbf{(Fig.~\ref{fig:fig5} (d))}} embeds modality‐specific inputs into a complex Hilbert space via quantum‐inspired mappings to capture long‐range dependencies. %in a single attention pass.
\textit{In parallel with $\texttt{MHDGA}$, the \texttt{MQIA} stream is designed to infuse quantum‐inspired frequency priors to guide the Lorentz information learning block. }
For each modality $i$, we leverage modality-specific frequency components $\psi_i$ to compute complex quantum states: $q_i \;=\;\psi_i\cdot\bigl(\eta_{\rm real}+j\,\eta_{\rm imag}\bigr)$, where \(\eta_{\rm real},\eta_{\rm imag}\) are learnable parameters.
By applying the Born rule \citep{Hall2013, NielsenChuang2010}, we compute the element-wise amplitudes of the quantum states: $|q_i|^2 = q_i\,q_i^*$, where \(q_i^*\) denotes the complex conjugate of $q_i$. 
We then project the real component of \(q_i\) into the Lorentz manifold using Eqs. \ref{eq:eq6}-\ref{eq:eq7}, allowing quantum states to be interpreted in hyperbolic spaces via mutual guidance \textcolor{black}{\textbf{(Fig.~\ref{fig:fig5} (c--d))}}, thus improving the learning of complex representational structural details.
Let \(\lambda_i = \|\texttt{MLE}(\psi_i)_{1:}\|\) be the hyperbolic Lorentz-norm of $\texttt{MLE}(\cdot)$ -- drawing on fractal-geometry principles, we define the quantum attention weights, followed by attention maps as:
%$A_i^{Q} = \texttt{MQIA}(\cdot)$ as:
%\vspace{-0.5cm} 
\begin{equation} \small
A_i^Q = \texttt{MQIA}(\psi_i) = \texttt{Softmax}\bigl(|q_i|^2 \,\bigl\|\,\texttt{MLE}(\psi_i)\bigr\|^{\,\!h-2}). 
%\vspace{-0.0cm} 
\label{eq:eq9}
\end{equation} 
%\vspace{-0.3cm}
where $h$ is a scalar hyperparameter controlling the strength of the quantum-inspired interaction.
The resulting $\texttt{MQIA}(\cdot)$ yields the bias \(\delta_i\) used in \texttt{LIL} sub-block. 
Through mutual guidance between \texttt{LIL} and \texttt{MQIA}, our \texttt{HySAM} module captures richer, geometry-aware hierarchical structural details across modalities.

\medskip
\noindent\textbf{B. Multimodal Attention Fusion Gating.} The \texttt{MAFG} \textcolor{black}{\textbf{(Fig.~\ref{fig:fig4} (b))}} block is designed that dynamically fuses geometry-aware attention maps from \texttt{MHDGA} mechanism with quantum-guided maps from \texttt{MQIA} mechanism. 
For each modality $i$, we use modality-specific learnable weights $c_i$, by adaptively weighting each modality's contribution to learn attention maps $A_{i}$. The \texttt{MFAG} block is formally defined as: 
%\vspace{-0.7cm} 
\begin{eqnarray} \small
%\small 
A_i & = & \texttt{MAFG}(\texttt{HQMGA}(x^{\prime}_i)), \nonumber \\
& = & \big( c_i \odot \texttt{MHDGA}\left( x^{\prime}_i\right) \big) + \big( c_i \odot \texttt{MQIA}\left( x^{\prime}_i \right) \big), \nonumber \\
& = & \big( c_i \odot  \sigma\bigl(L \times \texttt{LIL}(\psi_i) + P \times \texttt{PIL}(\psi_i) \bigr)  \big) + \big( c_i \odot \texttt{MQIA}(\psi_i) \big).
%\vspace{-0.0cm} 
\label{eq:eq10_1}
\end{eqnarray}
%\vspace{-0.3cm}
This adaptive fusion ensures robust integration of hyperbolic and quantum structural priors, enhancing cross-modal interactions and enabling the learning of complementary shared representations.

%\medskip
%\noindent \textit{Throughout this work, we use the term \emph{hybrid-space aware geometry} in a mechanistic sense:
%\texttt{HySAM} combines (i) a geometric branch implemented as exponential maps on negatively
%curved manifolds (Poincaré ball and Lorentz hyperboloid) and (ii) a quantum-inspired branch
%whose scores are squared amplitudes in a complex Hilbert space, fused through a
%probability-preserving, norm-stable gate; Appendix~\ref{sec:theory-physics1} formalizes this definition
%(Definition~\ref{def:phys-inspired-attn}) and proves that \texttt{HySAM} satisfies it (Theorem~\ref{thm:pcmfa-physics})}.

%\textbf{Learnable Late Fusion strategy} (Figure~\ref{fig:fig2}) denoted as $\texttt{LF}(\cdot)$, takes the shared representation $x_i^S$ from the \texttt{PCMFA} module and applies element‐wise modulation by trainable weights $\omega_i$, then fuses (concatenation) $\theta$, to capture a more expressive multimodal representation ($x^{S^{\prime}}_{j \in [1:m]}$): 
%$\texttt{LF}(x_i^S) \;=\; \theta\bigl(\omega_i \odot x_i^S\bigr)$. 

%\end{itemize}

\begin{algorithm}[H]
\renewcommand{\thealgorithm}{2}
\caption{\texttt{HyFuse}($x_{i-1},x_i$)}
\label{al:al2}
\begin{algorithmic}[1]
  \STATE \textbf{Input:} Heterogeneous modality inputs $x_{i-1},\,x_i$
  \STATE \textbf{Output:} Shared representations $\hat{x}^S_i,\,x^{S^\prime}_i$
  \STATE \textbf{Procedure:}
  \STATE \textit{/* Hybrid Geometry Aware Fusion (\texttt{HyFuse}) layer for multimodal input integration pipeline */}
  \STATE $\,x^{\prime}_{i-1},\,x^{\prime}_{i}\,\gets$ \texttt{EMRC}($x_{i-1},\,x_i$)  \quad  \quad   // Multi-scale cues learning Eq.~\ref{eq:eq9_1}.
  \STATE $\,x^{S}_{i-1},\,x^{S}_{i}\,\gets$  \texttt{HySAM}($x^{\prime}_{i-1},\,x^{\prime}_{i}$) \quad //  more expressive shared representation learning Eqs.~\ref{eq:eq1}–\ref{eq:eq10_1}.
  \STATE $\,x^{S}_{i-1},\,x^{S}_{i}\,\gets$ \texttt{HySAM}($x^{S}_{i-1},\,x^{S}_{i}$) \quad // Further enriches information.
  \STATE $x^{S^\prime}_{i}\gets$  \texttt{LLF}($x^{S}_{i-1},\,x^{S}_{i}$) \quad \quad   \quad // Handles missing modalities Eqs. \ref{13}--\ref{15}.
  \STATE $\hat{x}^{S}_{i}\gets$ Call \texttt{SIR}($x^{S}_{i}$) \quad \quad  \quad   \quad  \quad // Enriches robust shared features  Eq.~\ref{eq:eq10}.
  \STATE \textbf{Return:} $\hat{x}^{S}_{i},\,x^{S^\prime}_{i}$
\end{algorithmic}
\end{algorithm}

%%%%%%%%%
%%%%%%%%%%%%%%%%%%%%%%%
%\vspace{-0.2cm}
\begin{figure}[t]
    \centering
    \includegraphics[width=0.48\textwidth]{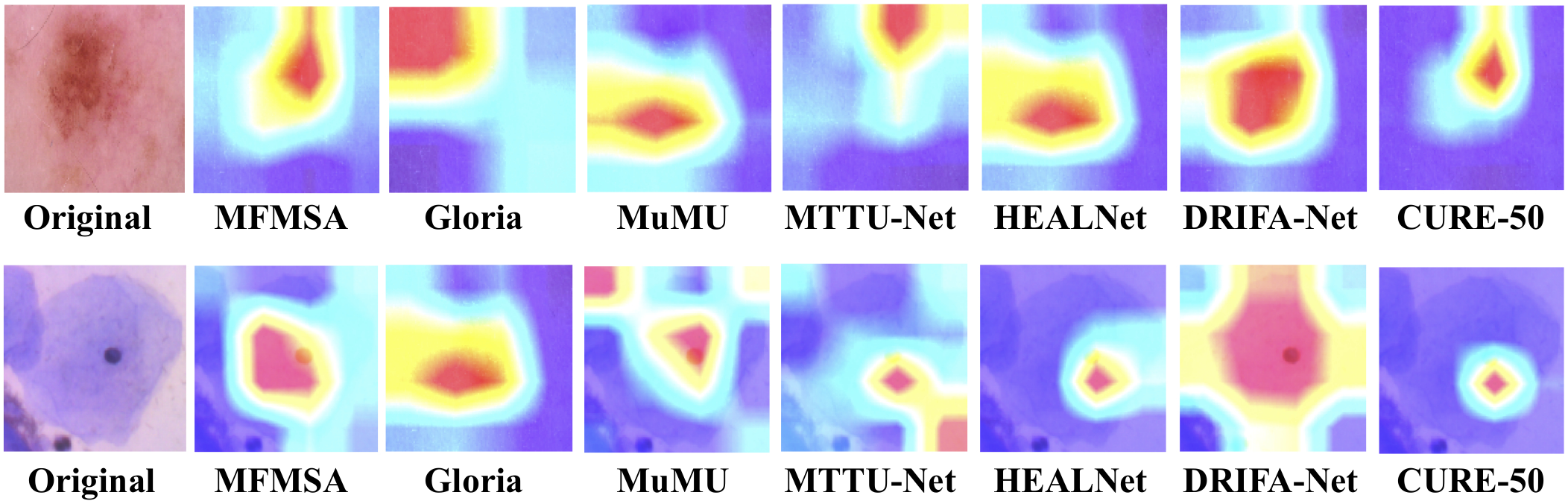}
    \vspace{-0.5cm}
      \caption{\small
Visual comparison of discriminative regions highlighted by our proposed \texttt{CURE} variant (e.g., \texttt{CURE-50}) and seven top-performing state-of-the-art methods using the \texttt{Grad-CAM} technique on two benchmark datasets: \texttt{HAM10000} (top row) and \texttt{SIPaKMeD} (bottom row)}
    \label{fig:fig8}
    \vspace{-0.3cm}
\end{figure}

\subsubsection{\textbf{Learnable Late Fusion}}\label{LF_strategy}
Given shared representations $\{x_i^{S}\}$ from the \texttt{HySAM} module, where each $x_i^{S}\!\in\!\mathbb{R}^{H\times W\times C}$, we design learnable a Late-Fusion (\texttt{LLF}) layer, which learns \emph{scalar}, content-aware gates that (i) adapt to the current sample, (ii) reduce exactly to zero whenever a modality is missing, and (iii) incur negligible overhead (one global-pooling operation and a small \texttt{MLP} per modality), thereby capturing robust shared features ($x^{S^{\prime}}_{j \in [1:m]}$). The overall LLF workflow is presented in the following steps:

\medskip

\noindent\textbf{(a) Channel context learning and mask-aware gating.}  
For each modality $i$, we summarize channel context via global average pooling (\texttt{GAP}) and map it to a scalar logit $z_i$ using a two-layer perceptron $h_i$ with \texttt{ReLU}: 
\begin{equation} \small
\label{13}
p_i = \mathrm{GAP}(x_i^{S}) \;\in\; \mathbb{R}^{C},
\quad 
z_i = h_i(p_i) \;\in\; \mathbb{R}.
\end{equation}

Let $mask \in \{0,1\}^{m}$ indicate modality availability. We mask missing modalities by multiplying them by a constant $o$ and then apply softmax to compute the attention weights, $\alpha_i$:
\begin{equation} \small
\label{14}
\tilde z_i = z_i + (1 - mask_i)\,o,
\quad
\alpha_i = \frac{\exp(\tilde z_i)}{\sum_{i=1}^{m} \exp(\tilde z_i)}.
\end{equation}
This yields $\alpha_i = 0$ when $mask_i = 0$ and $\sum_{i:mask_i = 1} \alpha_i = 1$.

\medskip
\noindent\textbf{(b) Gated fusion.}  
The resulting weights are used to fuse with the shared features $\{x^{S}_i\}_{i=1}^{m}$ corresponding to each modality $i$ and to capture robust shared features $x^{S'}_{j}$ ($j \in [1\!:\!m]$):
\begin{equation}
\small
\label{15}
x^{S'}_j
\;=\;
\texttt{LLF}\!\bigl(\{x_i^{S}\},\,mask_i\bigr)
\;=\;
\theta\!\bigl(\alpha_1 \ x_1^{S},\,\ldots,\,\alpha_m \ x_m^{S}\bigr)
\;\in\;
\mathbb{R}^{H \times W \times (mC)},
\end{equation}
where $\theta(\cdot)$ denotes concatenation.
At inference, at least one modality must be present.

%%%%%%%%%%%%%%

%It is important to note that, the Learnable Late Fusion ($\texttt{LF}(\cdot)$) approach is designed to leverage the resulting shared representations from \texttt{PCMFA} module along with trainable weights to modulate them to capture more expressive multimodal information ($x^{S^{\prime}}_{j \in [1:d]}$): $\texttt{LF}(x^{s}_i) = \theta( \omega \odot x^{s}_i )$.

%%%%%%%%%%%%%%%%%%%%%%%%%%%%%%%%%%
%\vspace{-0.1cm} 
\begin{table*}[tbp]
\centering
\setlength{\abovecaptionskip}{-0.0pt} % Reduce space above caption
%\captionsetup{aboveskip=-10pt, belowskip=-10pt}
  \centering
  %\captionsetup[table]{aboveskip=-4pt, belowskip=-4pt}
%\footnotesize
%\caption{\small Performance comparison of \texttt{EHPAL-Net} with SOTA methods (\texttt{M1–M14}) on heterogeneous medical datasets for multi-disease classification, patient survival and mortality predictions. Bold and underlined values indicate the best and second-best results, respectively. \texttt{EHPAL-Net} employs  \texttt{ResNet18}, \texttt{ResNet50}, \texttt{Inception-v3}, and \texttt{ShuffleNet} as base networks, denoted as \texttt{EDF-Net-18}, \texttt{EDF-Net-50}, \texttt{EDF-Net-IN}, and \texttt{EDF-Net-SN}.}
\caption{\small Performance comparison of \texttt{CURE} variants—\texttt{CURE-18}, \texttt{CURE-50}, \texttt{CURE-IN}, \texttt{CURE-V}, and \texttt{CURE-SN} 
%(rely on \texttt{ResNet18}, \texttt{ResNet50}, \texttt{Inception‑v3}, \texttt{ShuffleNet})
-- against \texttt{SOTA} models on unpaired heterogeneous modality aware datasets.
%for multi‑disease classification and survival/mortality prediction. 
\textbf{Bold} and \underline{underlined} indicate the best and the second-best results, respectively.}
\vspace{0.1cm}
%\textbf{(A)}
\label{tab:tab1}
\scalebox{0.8}{
\setlength{\tabcolsep}{6.75pt} % Adjust the column spacing (default is 6pt)
\begin{tabular}
{l|l|cc|cc|c|cc|cc|cc|cc|c|cc}
\toprule
\multirow{2}{*}{\textbf{Models}} & \multirow{1}{*}{\textbf{Datasets} $\rightarrow$} & \multicolumn{2}{c|}{\textbf{HAM10000}} & \multicolumn{2}{c|}{\textbf{SIPaKMeD}} & \textbf{BRCA} & \multicolumn{2}{c|}{\textbf{MORT}} & \multicolumn{2}{c|}{\textbf{ICD9}} & \multicolumn{2}{c|}{\textbf{PATHMNIST}} & \multicolumn{2}{c|}{\textbf{OrganAMNIST}}
& \textbf{UCEC} 
& \multicolumn{2}{c}{\textbf{Overall}}
\\ 
\cline{2-18}
  & Backbone & \multicolumn{1}{c}{ACC}  & AUC & \multicolumn{1}{c}{ACC}  & AUC & C-Index & \multicolumn{1}{c}{ACC}  & AUC & \multicolumn{1}{c}{ACC}  & AUC  & \multicolumn{1}{c}{ACC} &  \multicolumn{1}{c|}{AUC}
 & \multicolumn{1}{c}{ACC} &  \multicolumn{1}{c|}{AUC}  & C-Index & \multicolumn{1}{c}{\#P $\downarrow$}
& \multicolumn{1}{c}{\#F $\downarrow$}\\
 \midrule
 \rowcolor{gray!10}
 \multirow{1}{*}{POTTER} & ResNet18 & 91.35 &  91.72 & 92.40 & 92.66 & 55.41 & 85.18 & 86.55 
 & 67.20 & 90.17 & 91.46 & 99.58 
 & 96.06 & 99.85 & 50.28   
& {12} & {0.95} \\

\multirow{1}{*}{NAT} & Swin-T 
& 93.11 & 93.25 & 91.87 & 91.62 
& 54.34 & 86.25 & 88.29  
& 68.40 & 92.18 & 91.76 & 99.87 
& 95.72 & 99.60 & 50.47  
& 20 & 1.1 \\
\rowcolor{gray!10}
 \multirow{1}{*}{DDA-Net} & ResNet18
 & 92.83 & 92.14 & 92.39 & 92.61 
 & 56.46 & 87.16 & 88.68 
 & 66.85 & 91.22 & 92.24 & 99.65 
 & 95.73 & 99.75 & 50.80   
 & 12.1 & 1.12 \\

 \multirow{1}{*}{MFMSA} & ResNet50
 & 97.90 & 97.90 & 94.76 & 95.38 
 & 56.75 & 89.50 & 92.90  
 & 69.63 & 94.32 & 92.57 & 99.70 
 & 96.90 & 99.80 & 51.27  
 & 26.9 & 1.4 \\
 \rowcolor{gray!10}
\multirow{1}{*}{MSCAM} & PVT2-B2
& 97.55 & 97.71 & 94.25 & 95.05 
& 55.84 & 90.78 & 93.90 
& 70.10 & 94.84 & 93.18 & 99.75 
& 96.54 & \underline{99.90} & 50.13  
& 26.9 & 1.3 
\\
%\cline{1-27} %\cline{1-27}  
\midrule
%\hline

 \multirow{1}{*}{Gloria} & ResNet50
 & 93.75 & 94.58 & 94.32 & 94.50 
 & 58.63 & 89.15 & 92.28 
 & 65.0 & 88.60 & 92.41 & 99.58 
& 95.85 & 99.75 & 51.31 
& 30.8 & 1.54  \\
\rowcolor{gray!10}
\multirow{1}{*}{HAMLET} & ResNet50
& 93.25 & 93.20 & 92.84 & 93.32 
& 57.22 & 88.35 & 91.40 
& 67.12 & 90.58 & 92.20 & 99.80 
& 95.64 & 99.72 & 50.70  
& 57.3 & 3.52  \\

\multirow{1}{*}{MTTU-Net} & ResNet50 
& 97.45 & 97.18 & 91.90 & 92.56 
& 58.23 & 89.58 & 92.80  
& 68.5 & 92.81 & 92.48 & 99.50 
& 95.45 & 99.60 & 52.18  
& 38.1 & 6.8   \\
\rowcolor{gray!10}
\multirow{1}{*}{MuMu} & ResNet50
& 92.80 & 93.12 & 92.15 & 92.78 
& 57.92 & 88.74 & 91.85  
& 66.88 & 90.90 & 91.87 & 99.45 
& 95.64 & 99.80 & 51.27  
& 56.6 & 2.97  \\

\multirow{1}{*}{M$^3$Att} & Swin-B
& 95.80 & 95.94 & 92.32 & 92.88 
& 59.38 & 90.75 & 94.05  
& 68.10 & 92.48 & 93.20 & 99.90 
& 95.51 & 99.65 & 52.45 
& 183 & 12.14 \\
\rowcolor{gray!10}
\multirow{1}{*}{Perceiver} & ResNet50
& 92.42 & 92.57 & 91.55 & 91.70 
& 55.78 & 87.20 & 88.81  
& 71.30 & 96.52 & 93.45 & 99.80 
& 96.27 & \underline{99.90} & 61.47 
& 31.0 & 11.7 \\

\multirow{1}{*}{MOTCAT} & ResNet50
& 95.35 & 95.80 & 93.56 & 93.94 
& 57.35 & 90.27 & 93.78  
& 66.22 & 89.56 & 92.80 & 99.50 
& 95.10 & 99.58 & 52.57  
&  \underline{29.5} & 1.48 \\
\rowcolor{gray!10}
\multirow{1}{*}{HEALNet} & ResNet50
& 98.24 & 98.17 & 94.75 & 94.80 
& 57.10 & 91.24 & 94.30  
& 70.66 & 95.72 & 93.58 & 99.83 
& 96.12 & \underline{99.90} & 53.97  
& 31.6 & 3.84 \\

\multirow{1}{*}{DRIFA-Net} & ResNet18
& 98.33 & 98.51 & 95.58 & 95.75 
& 56.47 & 91.32 & 94.10  
& 70.28 & 95.10 & 93.45 & 99.75 
    & 96.45 & \underline{99.90} & 52.48  
& 53.8 & 4.83 \\
%\cline{1-27} \cline{1-27} 
\midrule 
\rowcolor{orange!10}
\multirow{1}{*}{\textbf{CURE-18}} & ResNet18 & 
\underline{99.75} & \textbf{99.99} & 
{95.97} & {99.67} & 
{63.34} &  
 {92.34} & {95.55} 
 & {73.64} & {98.24}  
 & 94.25 & 99.95 & {97.78} & \textbf{99.98}
 & 61.74  
 & 7.71 & \underline {0.59} \\

\rowcolor{purple!10}
\multirow{1}{*}{\textbf{CURE-50}} & ResNet50 & \textbf{99.81} & \textbf{99.99} & {96.32} & \underline{99.90} & {64.59} & 
\underline{92.52} & \underline{96.10} & \underline{76.53} & \underline{99.30} & 
\underline{94.63} & \textbf{99.98} & \underline{98.12} & \textbf{99.98} &
{62.20} & 
14.4 & 1.83  \\
\rowcolor{orange!10}
\multirow{1}{*}{\textbf{CURE-IN}} & Inception-v3
& 98.62 & \underline{98.81} 
& \underline{96.62} & \underline{99.90} 
& \underline{64.85} 
& 91.55 
& {94.81} & {74.20} & {98.85} &  {94.40} & \underline{99.96} & 
{97.15} & \underline{99.20} &
 \textbf{62.82} 
 & 13.7 & 2.82 \\

\rowcolor{purple!10}
\multirow{1}{*}{\textbf{CURE-V}} & ViT-Tiny & \underline{99.75} & \textbf{99.99} & \textbf{97.10} & \textbf{99.95} & \textbf{65.22} & 
\textbf{93.37} & \textbf{97.08} & \textbf{76.85} & \textbf{99.47} & 
\textbf{95.02} & 99.95 & \textbf{98.51} & \textbf{99.98} &
\underline{62.56} & 
10.8 & 1.25  \\
\rowcolor{orange!10}
\multirow{1}{*}{\textbf{CURE-SN}} & \text{ShuffleNet}
& 97.84 & 98.64 & 95.25 & 97.73 
& 62.14
& {90.37} 
& {93.64} 
& {71.41} 
& 97.10 
& {92.87} & {99.55} 
& 96.85 
& 99.0 &
{60.11} 
 & \textbf {3.1} & \textbf{0.29} \\
\bottomrule
\end{tabular}}
%\vspace{-0.8cm}
\end{table*}

%\vspace{-0.2cm}

%%%%%%%%%%%%%%%%%%%%%%%%%%
%%%%%%%%%

%%%%%%%%%%%%%%%%%%%%%%%%%%%%%%%%%%%
%\vspace{-0.4cm}
\subsubsection{\textbf{Shared Information Refinement with Intermediate Fusion}}\label{shared_sir} 

To promote modality-order-invariant robustness, at each \texttt{HyFuse} layer we refine the shared feature $x^{S}_{i}$ produced by \texttt{HySAM} via Shared Information Refinement (\texttt{SIR}), yielding $\hat{x}^{S}_{i}$.
\texttt{SIR} extends our \texttt{MHCF} block (\textcolor{black}{\textbf{App.~\ref{sir}; Fig.~\ref{fig:fig6}(C))}} by adding an extra $3{\times}3$ depthwise-convolution (\texttt{DWC}) branch, resulting in four heterogeneous branches with varying scales $(1, 3, 5)$: group point-wise convolution (\texttt{GPC}) with $1{\times}1$ kernel size, $\texttt{DWC}_{3{\times}3}$, $\texttt{DWC}_{5{\times}5}$, and dilated depthwise convolution $\texttt{DDC}_{3{\times}3}$ with dilation $r{=}2$ for enlarged context. 
We fuse the multi-scale outputs, apply channel shuffle $\texttt{CS}(\cdot)$ followed by \texttt{GPC} for efficient cross-branch mixing, and employ a residual connection with \texttt{GELU} to preserve salient cues.  
Finally, \texttt{SIR}’s cascaded structure increases the channel dimensions to align with each \texttt{HySAM} output, thereby learning enriched representations (\( \hat{x}^{S}_{i} = \texttt{SIR}(x_i^{S}) \)).

We further incorporate \emph{intermediate fusion} approach by aggregating the refined features $\hat{x}^{S}_{i}$ across \texttt{HyFuse} layers \textcolor{black}{\textbf{(Fig.~\ref{fig:fig3}(a))}} to encourage modality-order-invariant shared cues, and fusing them with the robust shared feature $x^{S^{\prime}}_{m}$ produced by \texttt{LLF} at the final \texttt{HyFuse} layer to obtain the modality-order-invariant robust shared features $x^{C}$ (Eq.~\ref{eq:eq10}), thereby improving \texttt{CURE}'s effectiveness. 

\begin{equation}\small
\texttt{SIR}(x^{S}_{i})
=
\texttt{G}\!\left(
x^{S}_{i}
+
\texttt{GPC}\!\left(
\texttt{CS}\!\left(
\theta\!\left(
\left\{\texttt{DC}_{sc,r}(x^{S}_{i})\right\}_{sc,r}
\right)
\right)
\right)
\right),
\label{eq:eq9}
\end{equation}

\begin{equation} \small
x^{C}
=
\theta\!\left(
x^{S^{\prime}}_{m},\;
\theta\!\left(
\left\{\texttt{SIR}(x^{S}_{i})\right\}_{i}
\right)
\right),
\label{eq:eq10}
\end{equation}
where $\texttt{DC}_{sc,r} \in \left\{
\texttt{GPC}_{1\times 1},
\texttt{DWC}_{3\times 3},
\texttt{DWC}_{5\times 5},
\texttt{DDC}_{3\times 3,\,r=2}
\right\}$.
%Details are provided in App.~\ref{sir}.

%%%%%%%%%%%%%%%%%%%%%%%%%%%%%%%%%%%%%%%%%%%

%\vspace{-0.05cm}
\subsection{\textbf{Heterogeneous Modality-specific Multitask Learning}}
%\vspace{-0.2cm}

The \texttt{HMML} phase \textcolor{black}{\textbf{(Fig.~\ref{fig:fig3} (b))}} is designed to leverage the modality-order-invariant robust shared representations \(x^C \) from the \texttt{MSIL} phase of \texttt{CURE} for multiple tasks across \( m \) modalities. It maps input \( \mathcal{X} \) to output predictions \( \mathcal{Y} \) using a loss function \( \mathcal{L}_{\texttt{HMML}} \), where \( \lambda_t^m \) controls the weight of each task-modality-specific loss \( \mathcal{L}_t^m \). The best model parameters \( \beta^* \) are found by minimizing \( \mathcal{L}_{\texttt{HMML}} \):

\vspace{-0.3cm}
%\begin{footnotesize}
\begin{equation} \small
\label{eq:eq11}
\mathcal{L}_{\texttt{HMML}} = \sum_{t=1}^{T} \sum_{m=1}^{M} \lambda_t^M \cdot \mathcal{L}_t^m \big( \mathcal{F}(x^C; \beta), \mathcal{Y} \big) \quad \text{and} \quad \beta^* = \arg \min_\beta  \left(\mathcal{L}_{\texttt{HMML}})\right.
\vspace{-0.1cm}
\end{equation}

%\end{footnotesize}
%\vspace{-0.1cm}
\noindent where $\beta$ signifies the \texttt{CURE} parameters.

%\medskip
%\noindent\textit{For a \textcolor{purple}{comprehensive theoretical analysis} that precisely formalizes (i) the \texttt{CURE} framework and (ii) how it addresses the three challenges (see Sec.~\ref{intro}) via the \texttt{HyFuse} layer---comprising the EMRC, HySAM, LLF, and SIR modules---\textcolor{purple}{please refer to App.~\ref{th} (\ref{ch1}--\ref{ch3})}. This analysis shows that the \texttt{HyFuse} layer enables efficient, modality-order-invariant shared-representation learning, thereby improving \texttt{CURE}'s performance--efficiency trade-off.}

%To ensure reliability in~\texttt{EHPAL-Net} framework, one can employ the Monte Carlo dropout strategy~\citep{gal2016dropout} to estimate uncertainty in the model's predictions.

%%%%%%%%%%%%%%%%%%%%%%%%%%%%%%%%%%%%%%%%%%%
%%%%%%%%%%%%%%%%%%%%%%%%%%%%%%%%%%%%%%%%%%%

%%%%%%%%%%%

\begin{table*}[t]
\centering
\captionsetup{aboveskip=2pt, belowskip=2pt}

\caption{\small
\textbf{(a)} Performance of the \texttt{CURE} family (\texttt{CURE-50}, \texttt{CURE-V}) compared with leading unimodal and multimodal baselines in paired regimes.
\textbf{(b)} Missing-modality robustness in paired regimes for \texttt{CURE-50}, trained with all modalities and evaluated under four test-time settings: (i) omics-only, (ii) whole-slide histopathology images (\texttt{WSI})-only, (iii) mixed (50\% of samples from each modality), and (iv) full (both modalities for all samples). \textbf{(c)} Fusion-scheme ablation: performance of alternative fusion strategies embedded in \texttt{CURE-18} across unpaired regimes.}
\label{tab:tab2}
\vspace{-2mm}

% --- Outer 1-row tabular prevents minipages from wrapping to the next line ---
\setlength{\tabcolsep}{0pt}
\begin{tabular}{@{}c@{\hspace{0.8em}}c@{\hspace{0.8em}}c@{}}

% ===================== (a) =====================
\begin{minipage}[t]{0.215\textwidth}
\centering
\vspace{0.1cm}
{\small\textbf{(a) Paired modalities.}}\\[-2pt]
\vspace{0.3cm}
\setlength{\tabcolsep}{4pt}
\renewcommand{\arraystretch}{1.05}

\begin{adjustbox}{max width=0.95\linewidth}
\begin{tabular}{l|c|c}
\hline
\textbf{Models} $\downarrow$ &
\textbf{BLCA} &
\textbf{KIRP}  \\
\toprule

MFMSA &
57.2{\scriptsize$\pm1.06$} &
64.4{\scriptsize$\pm1.28$}   \\[1pt]
\midrule

Perceiver\textsuperscript{*} &
54.7{\scriptsize$\pm1.61$} &
69.2{\scriptsize$\pm3.57$}  \\[1pt]
CA-MLIF\textsuperscript{*} &
70.7{\scriptsize$\pm2.25$} &
73.2{\scriptsize$\pm1.38$}  \\[1pt]

HEALNet &
71.4{\scriptsize$\pm2.73$} &
81.2{\scriptsize$\pm2.87$}  \\[1.5pt]
MMLego &
73.4{\scriptsize$\pm1.93$} &
86.3{\scriptsize$\pm1.52$}  \\
\midrule
\rowcolor{orange!10}
\textbf{CURE-50} &
{\underline{75.1}{\scriptsize$\pm1.15$}} &
{\underline{87.2}{\scriptsize$\pm1.03$}}  \\[1.5pt]
\rowcolor{purple!10}
\textbf{CURE-V} &
\textbf{75.9}{\scriptsize$\pm0.82$} &
\textbf{87.8}{\scriptsize$\pm0.57$}  \\[1.5pt]
\bottomrule
\end{tabular}
\end{adjustbox}
\end{minipage}
&
% ===================== (b) =====================
\begin{minipage}[t]{0.37\textwidth}
\vspace{0.1cm}\centering
{\small\textbf{(b) Missing-modality evaluation.}}\\[-2.5pt]
\vspace{0.3cm}
\setlength{\tabcolsep}{2pt}
\renewcommand{\arraystretch}{1.02}

\begin{adjustbox}{max width=\linewidth}
\begin{tabular}{@{}l|cc|cc|cc|cc@{}}
\toprule
\multirow{2}{*}{\textbf{Models}} &
\multicolumn{2}{c|}{\textbf{Omics-only}} &
\multicolumn{2}{c|}{\textbf{WSI-only}} &
\multicolumn{2}{c|}{\textbf{50\% of Both}} &
\multicolumn{2}{c}{\textbf{100\% of Both}}  \\[1.5pt]
\cline{2-9}
 & \textbf{BLCA} & \textbf{KIRP} & \textbf{BLCA} & \textbf{KIRP} & \textbf{BLCA} & \textbf{KIRP} & \textbf{BLCA} & \textbf{KIRP}  \\
\midrule

MFMSA & 59.7 & 77.1 & 49.5 & 51.8 & 54.2 & 64.4 & 57.2 & 64.4 \\[3.5pt]

Perceiver & 68.6 & 83.6  & 53.2 & 71.6 & 60.1 & 66.9 & 54.7 & 69.2 \\[3.5pt]
CA-MLIF & 64.5 & 82.8 & 51.2 & 59.5 & 55.1 & 69.2 & 70.7 & 73.2 \\[3.5pt]

HealNet & 61.8 & 77.3 & 50.1 & 52.6 & 61.2 & 71.4 & 71.4 & 81.2 \\[3.5pt]
MMLego & 68.1 & 84.0 & 56.8 & 63.0 & 62.9 & 75.6 & 73.4 & 86.3 \\[3pt]
\midrule
\rowcolor{orange!10}
\textbf{CURE-50} & \textbf{68.7} & 84.9 & \textbf{57.9} & 66.3 & \textbf{64.9} & 76.9 & 75.1 & 87.2 \\[3.5pt]
\bottomrule
\end{tabular}
\end{adjustbox}
\end{minipage}
&
% ===================== (c) =====================
\begin{minipage}[t]{0.35\textwidth}
\vspace{0.1cm}\centering
{\small\textbf{(c) Fusion-scheme ablations.}}\\[-2pt]
\vspace{0.3cm}
\setlength{\tabcolsep}{2pt}
\renewcommand{\arraystretch}{1.02}

\begin{adjustbox}{max width=\linewidth}
\begin{tabular}{@{}l|cc|cc|c|cc@{}}
\toprule
{} & \multicolumn{2}{c|}{\textbf{HAM10000}} & \multicolumn{2}{c|}{\textbf{SIPaKMeD}} & \textbf{BRCA} & \multicolumn{2}{c}{\textbf{MORT}} \\
\midrule
\textbf{Model} & {ACC} & {AUC} & {ACC} & {AUC} & {C-Index} & {ACC} & {AUC} \\
\midrule

Early & 94.9 & 95.4 & 92.1 & 95.3 & 57.2 & 86.3 & 88.8 \\[3pt]
Intermediate & 98.4 & 98.3 & 94.9 & 97.9 & 59.8 & 89.7 & 92.3 \\[3pt]

Late & 97.8 & 98.1 & 94.4 & 97.3 & 57.9 & 88.9 & 93.8 \\[3.5pt]

Hybrid Early & 98.6 & 98.7 & 95.6 & 98.3 & 60.5 & 91.1 & 94.3 \\[3.5pt]
\midrule
\rowcolor{orange!10}
\textbf{HyFuse} & \textbf{99.8} & \textbf{99.9} & \textbf{95.97} & \textbf{99.7} & \textbf{63.3} & \textbf{92.3} & \textbf{95.6} \\[3.5pt]
\bottomrule
\end{tabular}
\end{adjustbox}
\end{minipage}

\end{tabular}

\vspace{-2mm}
\end{table*}

%\vspace{-0.2cm}
\section{Experimental Analysis and Results} \label{experiments_analysis_results}
%\vspace{-0.2cm}
\noindent
\textbf{Datasets.} We evaluate \texttt{CURE} on 16 heterogeneous public medical datasets in both unpaired\footnote{\emph{Unpaired}: datasets come from different cohorts/institutions and have no subject-level alignment across modalities; in our MSIL setting, each dataset is treated as an independent modality stream/task.} and paired\footnote{\emph{Paired}: multiple data types are available for the same patient/sample (e.g., Whole-slide histopathology images (WSI) and multi-omics profiles).} settings, grouped as follows:
(1) \textbf{Unpaired imaging} (\textbf{\textcolor{blue}{\texttt{D1}--\texttt{D7}}}):
\texttt{HAM10000}~\citep{tschandl2018ham10000},
\texttt{SIPaKMeD}~\citep{plissiti2018sipakmed},
\texttt{PathMNIST} and \texttt{OrganAMNIST} (\texttt{MedMNIST})~\citep{yang2023medmnist},
\texttt{SARS-CoV-2 CT-Scan}~\citep{angelov2020towards},
\texttt{CNMC-2019}~\citep{mourya2019all},
\texttt{Chest X-ray Pneumonia}~\citep{kermany2018labeled};
(2) \textbf{Unpaired multi-omics} (\textbf{\textcolor{blue}{\texttt{D8}--\texttt{D10}}}): TCGA \texttt{BRCA}~\citep{lingle2016radiology}, \texttt{UCEC}, and \texttt{GBMLGG};
(3) \textbf{Unpaired clinical/EHR \& time-series} (\textbf{\textcolor{blue}{\texttt{D11}--\texttt{D13}}}):
\texttt{MIMIC-III}~\citep{johnson2016mimic},
\texttt{MHEALTH}~\citep{Banos2014},
\texttt{UCI-HAR}~\citep{Anguita2013};
(4) \textbf{Paired imaging} (\textbf{\textcolor{blue}{\texttt{D14}}}): \texttt{BraTS-2021}~\citep{baid2021rsna};
(5) \textbf{Paired WSI + multi-omics} (\textbf{\textcolor{blue}{\texttt{D15}--\texttt{D16}}}): TCGA \texttt{KIRP}~\citep{linehan2016radiology} and \texttt{BLCA}~\citep{cancer2014comprehensive}.
Images are resized to $128\times128\times3$ and split $80/10/10$ into train/val/test (where applicable), with standard augmentations.
All experiments use five random seeds; we report the mean and (where applicable) standard deviation.

\medskip
\noindent
\textbf{Baselines.} We compare \texttt{CURE} against state‑of‑the‑art (\texttt{SOTA}) single‑modal learning and multimodal fusion learning (\texttt{MFL}) methods. For \texttt{D1}–\texttt{D15} datasets:  
(1) \textcolor{black}{\emph{Single‑modal learning models}} (\textcolor{blue}{\textbf{\texttt{M1–M5}}}): \texttt{POTTER} \citep{zheng2023potter}, \texttt{NAT} \citep{hassani2023neighborhood}, \texttt{DDA‑Net} \citep{cui2023dual}, \texttt{MFMSA} \citep{nam2024modality}, \texttt{MSCAM}~\citep{rahman2024emcad};  
(2) \textcolor{black}{\emph{Early fusion \texttt{MFL} model}} (\textcolor{blue}{\textbf{\texttt{M6}}}): Perceiver \citep{jaegle2021perceiver};  
(3) \textcolor{black}{\emph{Late fusion \texttt{MFL} models}} (\textcolor{blue}{\textbf{\texttt{M7–M11}}}): \texttt{Gloria} \citep{huang2021gloria}, \texttt{HAMLET} \citep{islam2020hamlet}, \texttt{MuMu} \citep{islam2022mumu}, \texttt{MTTU‑Net}~\citep{cheng2022fully}, \texttt{M$^3$Att} \citep{liu2023multi};  
(4) \textcolor{black}{\emph{Intermediate fusion \texttt{MFL} models}} (\textcolor{blue}{\textbf{\texttt{M12–M15}}}): \texttt{MOTCAT} \citep{xu2023multimodal}, \texttt{DRIFA‑Net} \citep{dhar2024multimodal}, \texttt{CA-MLIF}~\citep{an2025mlif}, and \texttt{MMLego} \citep{hemker2025multimodallegomodelmerging}; and 
(5) \textcolor{black}{\emph{Hybrid early fusion \texttt{MFL} model}} (\textcolor{blue}{\textbf{\texttt{M16}}}): \texttt{HEALNet} \citep{hemker2024healnet}.   
Our \texttt{CURE} is instantiated with \textcolor{black}{five backbones}—\texttt{ResNet18}, \texttt{ResNet50}~\citep{he2016deep}, \texttt{Inception‑v3}~\citep{szegedy2016rethinking}, \texttt{ViT-Tiny} \citep{steiner2021train} and \texttt{ShuffleNet}~\citep{zhang2018shufflenet} -- yielding variants \texttt{CURE‑18}, \texttt{CURE‑50}, \texttt{CURE‑IN}, \texttt{CURE‑V}, and \texttt{CURE‑SN}.

\medskip
\noindent
\textbf{Evaluation Metrics and Training Details.} We report the following metrics: accuracy (\texttt{ACC}), \texttt{AUC}, concordance index (\texttt{C-Index}), \textcolor{black}{\textbf{number of parameters (\texttt{\#P}) in millions, and floating‐point operations (\texttt{\#F}) in \texttt{GFLOPs}}}. 
All models are trained for $200$ epochs on a single \texttt{NVIDIA A100} 80GB \texttt{GPU} running on an Ubuntu machine.
(1) Loss: Negative log‑likelihood (\texttt{NLL}) for survival prediction (WSI and multi‑omics), and cross‑entropy for all classification tasks (imaging, \texttt{EHR}).
(2) Optimizer: Adam with initial learning rate \(1\times10^{-3}\).
(3) Scheduler: ReduceLROnPlateau down to \(1\times10^{-6}\).
\emph{\textbf{Further details on the data augmentation and implementations are provided in App.~\ref{data_augmentation} and ~\ref{implementataion_details}}}.

%%%%%%%%%%%%%%%%%%%%%%%

% Requires: \usepackage[table]{xcolor}

%%%%%%%%%%%%%%%%%%%%%%%%%%%%%%%%%%

%\vspace{-0.3cm}
\subsection{Performance Comparisons} \label{Performance_Comparisons}
%\vspace{-0.2cm} 

\medskip
\noindent\textbf{\textit{Unpaired heterogeneous modalities.}} \textbf{Table \ref{tab:tab1}} summarizes results across heterogeneous unpaired modality streams spanning imaging, multi-omics, and EHR for diverse downstream tasks (e.g., multi-disease classification, patient survival, and mortality prediction tasks). Across these tasks, the CURE family establishes a new state of the art, consistently outperforming both leading unimodal learners (e.g., MFMSA, MSCAM) and multimodal fusion baselines (e.g., HEALNet, DRIFA-Net) by up to $\approx 5.8\%$ over the strongest competing method. Importantly, these gains come at substantially lower cost: CURE variants lie in the 3.1–14.4M parameter range with 0.29–2.82 GFLOPs, yielding up to $\approx98\%$ reductions in parameters/FLOPs relative to computationally intensive prior multimodal fusion baselines (e.g., M$^{3}$Att), while still providing large savings ($\approx86$–$88\%$) compared to competitive intermediate-fusion paradigms such as DRIFA-Net (Table \ref{tab:tab1}). This positions CURE on the leading edge of the performance–efficiency frontier for heterogeneous, unpaired medical data. 

\medskip
\noindent\textbf{\textit{Paired multimodal benchmarks.}} On paired modalities (WSI and omics) for survival prediction on BLCA and KIRP (\textbf{Table \ref{tab:tab2}(a)}), CURE-50 and CURE-V again achieve the best performance (C-index), improving over the strongest prior multimodal fusion baseline by up to $\approx 2.5\%$ and by substantially larger margins over unimodal baselines. Together, Tables \ref{tab:tab1} and \ref{tab:tab2}(a) show that CURE remains state-of-the-art in both unpaired and paired regimes, rather than being specialized to any single modality configuration.

%%%%%%%%

\begin{table*}[tbp]
\centering
\caption{\small Ablation studies of \texttt{CURE-18} on heterogeneous unpaired-modality benchmarks. \textbf{Left:} module-wise ablations measuring the progressive contribution of each component. \textbf{Right:} HySAM replacement ablations, where HySAM is substituted with alternative attention designs while keeping the remaining pipeline fixed.}
\label{tab:tab3}
% ------------------- Left table: Module-wise ablations -------------------
\begin{minipage}[t]{0.51\linewidth}
\centering
\vspace{-0.2cm}
%\footnotesize
%\captionof{table}{Module-wise ablations: progressive effect of each \texttt{CURE-18} component on diverse unpaired modalities benchmarks (i.e., HAM10000, SIPaKMeD, BRCA, and MORT).}
\textbf{\small (a) Module-wise ablations.}\\[4pt]
%\vspace{-0.2cm}
\scalebox{0.79}{
\setlength{\tabcolsep}{2.5pt}
\begin{tabular}{@{}cccc|cc|cc|c|cc|cc@{}}
\hline
\multicolumn{4}{c|}{\textbf{Components}} & \multicolumn{2}{c|}{\textbf{HAM10000}} & \multicolumn{2}{c|}{\textbf{SIPaKMeD}} & \textbf{BRCA} &  \multicolumn{2}{c|}{\textbf{MORT}} & \multicolumn{2}{c}{} \\ 
\cline{1-13}
EMRC & HySAM & LLF & SIR & ACC & AUC & ACC & AUC & C-Index & ACC & AUC & \#P & \#F \\ 
\hline
\rowcolor{gray!10}
$\times$ & $\times$ & $\times$ & $\times$ & 89.3 & 89.3 & 87.7 & 87.9 & 53.6 & 74.5 & 76.9 & {11.2} & {\underline{0.59}} \\[1pt]

$\checkmark$ & $\times$ & $\times$ & $\times$ & 92.7 & 92.8  & 91.8 & 92.1 & 55.4 & 77.9 & 78.2 & {\textbf{5.53}} & {\textbf{0.38}} \\[1pt]
\rowcolor{gray!10}
$\times$ & $\checkmark$ & $\times$ & $\times$ & 98.3 & 98.3 & 95.3 & 98.1 & 61.9 & 90.4 & 93.8 & 13.7 & 1.12 \\[1pt]

$\times$ & $\times$ & $\checkmark$ & $\times$ & 91.5 & 91.8 & 91.1 & 91.2 & 54.9 & 79.6 &  79.6 & 12.2 & 0.65 \\[1pt]
\rowcolor{gray!10}
$\times$ & $\times$ & $\times$ & $\checkmark$ & 93.6 & 93.7 & 92.2 & 92.7 & {57.2} & {81.8} & 82.2 & 13.7 & 0.89 \\[1.25pt]

$\checkmark$ & $\times$ & $\checkmark$ & $\checkmark$ & 94.5 & 94.5 & 93.6 & 93.9 & 58.7 & 85.1 & 85.7 & {\underline{6.82}} & 0.47 \\[1.25pt]
\rowcolor{gray!10}
$\checkmark$ & $\checkmark$ & $\checkmark$ & $\times$ & 99.2 & 99.2 & 95.6 & 98.9 & 62.3 & 91.1 & 94.0 & 7.53 & 0.55 \\[1.25pt]

$\times$ & $\checkmark$ & $\checkmark$ & $\checkmark$ & {\underline{99.4}} & {\underline{99.4}} & {\underline{95.7}} & {\underline{99.1}} & {\underline{62.7}} & 91.5 & 94.4 & 14.7 & 1.45 \\[1.25pt]
\rowcolor{orange!10}
$\checkmark$ & $\checkmark$ & $\checkmark$ & $\checkmark$ & 99.8 & 99.9 & 96.0 & 99.7 & {\textbf{63.3}} & {\textbf{92.3}} & {\textbf{95.6}} & 7.71 & 0.59 \\[1.25pt]
\hline
\end{tabular}
}
\end{minipage}
\hfill
% ------------------- Right table: Attention-scheme ablations -------------------
\begin{minipage}[t]{0.47\linewidth}
\vspace{-0.2cm}
\centering
%\footnotesize
%\captionof{table}{HySAM replacement ablations: performance of alternative attentions embedded in \texttt{CURE-18} on heterogeneous unpaired modalities.}
\textbf{\small (b) HySAM replacement ablations.}\\[4pt]
%\vspace{-0.2cm}
%\label{tab:tab5}
\scalebox{0.79}{
\setlength{\tabcolsep}{5pt}
\begin{tabular}{@{}l|cc|cc|c|cc|cc@{}}
\hline
\multirow{2}{*}{\textbf{Approaches}}   & \multicolumn{2}{c|}{\textbf{HAM10000}} & \multicolumn{2}{c|}{\textbf{SIPaKMeD}} & \textbf{BRCA} &  \multicolumn{2}{c|}{\textbf{MORT}}  & \multicolumn{2}{c}{} \\ 
\cline{2-10}
 & ACC & AUC & ACC & AUC & C-Index & ACC & AUC & \#P & \#F \\ 
\hline

CBAM~\cite{woo2018cbam} & 95.7 & 95.9 & 94.3 & 95.9 & 60.5 & 88.7 & 89.1 & 7.15 & 0.57 \\\rowcolor{gray!10}
SA~\cite{vaswani2017attention} & 96.3 & 96.8 & 94.7 & 97.4 & 61.2 & 90.1 &  90.8 & 8.45 & 0.85 \\
CA~\cite{chen2021crossvit} & 94.6 & 94.6 & 93.9 & 94.4 & 59.3 & 88.4 & 88.7 & {\textbf{6.95}} & {\textbf{0.49}} \\
\rowcolor{gray!10}
HSA~\cite{hou2024hsa} & 97.6 & 97.8 &  95.1 & 97.9 & 61.9 & 91.2 & 91.9 & 7.08 & 0.52 \\
QSAN~\cite{shi2022qsan} & 96.8 & 96.8 & 94.9 & 97.2 & 61.2 & 91.6 & 92.5 & 7.25 & 0.55 \\
\rowcolor{gray!10}
PIL (Ours)  & 96.4 & 96.4 & 94.4 & 95.1 & 59.4 & 90.9 & 92.2 & 7.49 & 0.51 \\
LIL (Ours) & 95.3 & 95.5 &  94.1 & 94.7 & 58.7 & 88.7 & 90.2 & 7.24 & 0.51 \\
\rowcolor{orange!10}
MHDGA (Ours) & 98.1 & 98.3 & 95.2 & 98.1 & 62.1 & 91.4 & 93.9 &{\underline{7.53}} &{\underline{0.55}} \\
\rowcolor{cyan!10}
MQIA (Ours) & 96.2 & 96.2 & 94.5 & 95.7 & 60.5 & 91.2 &  92.1 & 7.08 & 0.51 \\
\rowcolor{purple!10}
\textbf{HySAM} (Ours) & {\textbf{99.8}} & {\textbf{99.9}} &  {\textbf{96.0}} & {\textbf{99.7}} &{\textbf{63.3}} & 92.3 & 95.6 & 7.71 & 0.59 \\
\hline
\end{tabular}
}
\end{minipage}

\end{table*}

%%%%%%%%%%%

%%%%%%%%%%%
\medskip
 \textbf{\textit{\textcolor{blue}{To tackle Challenge~\ref{C1}}}} (\textbf{performance}), HyFuse consistently improves performance in both unpaired and paired regimes by learning robust shared features, yielding gains of up to $\approx$5.8 points on heterogeneous unpaired streams (\textbf{Table~\ref{tab:tab1}}) and up to $\approx$2.5 C-index points in paired regimes over prior leading baselines (\textbf{Table~\ref{tab:tab2}(a)}). Specifically, EMRC enriches each modality with discriminative multi-scale cues, which in turn improves complementary representation learning, while HySAM preserves coarse-to-fine structural cues, thereby enabling richer cross-modal relationships across heterogeneous sources. In addition, LLF gating and SIR, respectively, progressively aggregate, filter, and refine shared information across HyFuse layers, yielding robust shared representations that remain stable under modality heterogeneity and modality ordering. In contrast, prior early-, late-, and intermediate-fusion-aware multimodal fusion learning (MFL) designs often discard salient coarse-to-fine structural cues, limiting their ability to model complex cross-modal dependencies and leading to less expressive shared representations---consistent with the observed margins over existing multimodal fusion baselines. 

 \medskip
\textbf{\textit{\textcolor{blue}{To address Challenge~2}}} (\textbf{efficient design}), CURE's efficiency is primarily driven by the lightweight HyFuse implementation---especially the EMRC and SIR modules---together with a single-pass cascaded fusion pipeline that avoids multi-encoder and iterative-attention designs. While HySAM is introduced mainly to improve representation quality (\emph{\textbf{\textcolor{blue}{Challenge~1}}}), its design is also computationally moderate compared with large attention-matrix fusion (e.g., DRIFA-Net \citep{dhar2024multimodal}): it relies on computationally efficient attention mixing rather than high-dimensional token-token attention (e.g., self-attention \citep{vaswani2017attention}), helping CURE maintain the strong performance--cost trade-off observed in Table~\ref{tab:tab1} \emph{(e.g., up to $\approx$98.3\% fewer parameters and $\approx$97.6\% fewer FLOPs versus computationally intensive fusion baselines such as M$^{3}$Att \citep{liu2023multi}).}

%\vspace{-0.5cm}

\textbf{\textit{\textcolor{blue}{To deal with Challenge~3}}} (\textbf{generalization}), \textbf{Tables~\ref{tab:tab1} and \ref{tab:tab2}(a)} together provide evidence of generalization across heterogeneous medical modalities and learning regimes: CURE remains consistently strong across diverse unpaired modality streams (imaging, EHR, multi-omics) and also achieves superior performance in paired regimes without redesigning the architecture---supporting broad applicability across heterogeneous modality scenarios \emph{(maintaining gains of up to $\approx$5.8 points in unpaired settings and up to $\approx$2.5 C-index points in paired settings; Tables~\ref{tab:tab1} and \ref{tab:tab2}(a))} over leading prior baselines.

\medskip
\noindent\textit{\textbf{Effectiveness on missing–modality evaluation.}} \textbf{Table~\ref{tab:tab2}(b)} shows that CURE-50, trained with both WSI and omics, remains robust when modalities are missing at inference. It leads in all BLCA settings and in three out of four KIRP settings, and remains competitive in the remaining WSI-only regime \emph{(leading by $\approx$0.1--2.0 C-index points on BLCA and by up to $\approx$1.3 points on KIRP in the regimes where it performs best; and still exceeding MMLego by $\approx$3.3 points on KIRP under WSI-only).} This robustness is consistent with the LLF gating module, which can rely on whichever modality is available, and with the progressive refinement of shared information provided by SIR, which stabilizes the learned representation under modality dropouts.

%%%%%%%%%%%%%

\begin{table}[t]
\centering
\setlength{\tabcolsep}{4pt}
\renewcommand{\arraystretch}{1.05}
\caption{\small Performance comparison across diverse resolution scaling on representative imaging (HAM10000, SIPaKMeD),
multi-omics (BRCA), and EHR (MORT) tasks. Performance uses Acc. (\%) for
HAM10000/SIPaKMeD/MORT and C-index (\%) for BRCA. The $128\times128$ row is
reported in Table~\ref{tab:tab1}.}
\label{tab:res_scaling}
\vspace{-0.3cm}
\scalebox{0.8}{
\begin{tabular}{c l c c c c c c}
\toprule
\textbf{Input} & \textbf{Model} & \textbf{\#P (M)$\downarrow$} & \textbf{\#F$\downarrow$} &
\textbf{HAM10000} & \textbf{SIPaKMeD} & \textbf{BRCA} & \textbf{MORT}\\
\midrule
\rowcolor{orange!10}
\cellcolor{white}\multirow{3}{*}{128$\times$128}
 & 
 \textbf{CURE-V}     & 10.8 & 1.25  & \textbf{99.75} & \textbf{97.10} & \textbf{65.22} & \textbf{93.37}\\
 & HEALNet    & 31.6 & 3.84  & 98.24 & 94.75 & 57.10 & 91.24\\
 & DRIFA-Net  & 53.8 & 4.83  & 98.33 & 95.58 & 56.47 & 91.32\\
\midrule
\rowcolor{orange!10}
\cellcolor{white}\multirow{3}{*}{224$\times$224}
 & 
 \textbf{CURE-V}     & 10.8 & 4.24  & \textbf{99.78} & \textbf{97.35} & \textbf{65.30} & \textbf{93.45}\\
 & HEALNet    & 31.6 & 11.76 & 98.45 & 95.10 & 57.20 & 91.32\\
 & DRIFA-Net  & 53.8 & 14.79 & 98.62 & 96.07 & 56.58 & 91.51\\
\midrule
\rowcolor{orange!10}
\cellcolor{white}\multirow{3}{*}{256$\times$256}
 & 
 \textbf{CURE-V}     & 10.8 & 5.79  & \textbf{99.79} & \textbf{97.45} & \textbf{65.35} & \textbf{93.82}\\
 & HEALNet    & 31.6 & 15.36 & 98.50 & 95.21 & 57.26 & 91.43\\
 & DRIFA-Net  & 53.8 & 19.32 & 98.79 & 96.21 & 56.74 & 91.82\\
\midrule
\rowcolor{orange!10}
\cellcolor{white}\multirow{3}{*}{384$\times$384}
 & 
 \textbf{CURE-V}     & 10.8 & 15.99 & \textbf{99.80} & \textbf{97.57} & \textbf{65.46} & \textbf{94.11}\\
 & HEALNet    & 31.6 & 34.56 & 98.58 & 95.30 & 57.35 & 91.43\\
 & DRIFA-Net  & 53.8 & 43.47 & 98.90 & 96.35 & 56.93 & 92.05\\
\midrule
\rowcolor{orange!10}
\cellcolor{white}\multirow{3}{*}{512$\times$512}
 & 
 \textbf{CURE-V}     & 10.8 & 35.79 & \textbf{99.80} & \textbf{97.60} & \textbf{66.28} & \textbf{94.50}\\
 & HEALNet    & 31.6 & 61.44 & 98.58 & 95.45 & 57.35 & 91.43\\
 & DRIFA-Net  & 53.8 & 77.28 & 99.0 & 96.40 & 57.23 & 92.17\\
\bottomrule
\end{tabular}}
\end{table}

%%%%%%

\medskip
\noindent\textit{\textbf{Resolution-scaling evaluation.}}
We evaluate accuracy--efficiency trade-offs from \(128^2\) to \(512^2\)
resolution on representative imaging (\texttt{HAM10000}, \texttt{SIPaKMeD}),
multi-omics (\texttt{BRCA}), and \texttt{EHR} (\texttt{MORT}) tasks
(\textbf{Table~\ref{tab:res_scaling}}). Across all resolutions, \texttt{CURE-V}
remains the best-performing model while preserving a lower cost than
\texttt{HEALNet} and \texttt{DRIFA-Net}. Gains are modest on 
imaging benchmarks (\(\approx0.8\)--\(1.5\) points), but remain substantial
on structured modalities, with \(\approx8\)--\(9\) C-index points on
\texttt{BRCA} and \(\approx2\)--\(2.3\) points on \texttt{MORT}. Although
FLOPs increase with resolution, \texttt{CURE-V} keeps a fixed 10.8M-parameter
budget and consistently lower computation, e.g., \(1.25\) vs. \(3.84\)--\(4.83\)
GFLOPs at \(128^2\), and \(35.79\) vs. \(61.44\)--\(77.28\) GFLOPs at
\(512^2\). These results show that \texttt{CURE-V} scales robustly with
resolution, with diminishing returns on imaging tasks but clear
benefits on harder structured modalities.

\medskip
\noindent\textit{\textbf{Qualitative Analysis.}}
\textbf{Figure~\ref{fig:fig8}} compares discriminative regions identified by
\texttt{CURE} and competing baselines. \texttt{POTTER} highlights broad,
weakly localized regions, while \texttt{MuMu} reduces background activation
but still attends to non-informative areas. \texttt{MFMSA} and \texttt{HEALNet}
better localize the target structure, although spurious surrounding
activations remain. \texttt{DRIFA-Net} produces sharper boundaries but can
retain edge-level artifacts. In contrast, \texttt{CURE-18} most consistently
suppresses irrelevant background cues and concentrates activation on
anatomically meaningful regions, yielding the cleanest separation between
discriminative and non-discriminative evidence.
\textbf{\emph{Additional experimental analyses are provided in App.~\ref{Additional_Experiments}}.}

%%%%%%%%%%%%%%%%

%Unlike cascaded attention methods like \texttt{MFMSA}, \texttt{MSCAM}, and \texttt{DRIFA-Net}, the parallel attention design in ESFIA helps prevent gradual information loss, directly \textit{\textcolor{blue}{addressing Challenge 2}}. This is because cascaded designs are inherently \textcolor{purple}{limited in preserving holistic information} due to their \textcolor{purple}{sequential nature}, which tackles different aspects independently without \textcolor{purple}{lacking joint optimization}, ultimately \textcolor{purple}{constraining the richness of learned representations}. \texttt{EDF-Net} overcomes this limitation through parallel spatial-frequency fusion, facilitating richer shared representation learning. 
%While \texttt{DRIFA-Net} delivers competitive performance, its \textcolor{purple}{high computational cost limits its practicality in AI-driven healthcare}. In contrast, \texttt{EDF-Net} offers a \textcolor{purple}{more efficient architecture}, making it better suited for resource-constrained environments and enhancing generalizability across diverse medical imaging modalities (ref. Tables~\ref{tab:tab1}--\ref{tab:tab2}).

%\vspace{-0.3cm} 
\subsection{Ablation Study} \label{Ablation}
%\vspace{-0.2cm} 
%\medskip
\noindent \emph{\textbf{Fusion-scheme analysis.}} We conduct fusion-scheme ablations by replacing HyFuse with diverse fusion schemes (\textbf{Table~\ref{tab:tab2}(c)}), which isolates our CURE multimodal fusion design choices within the same backbone and shows that HyFuse consistently outperforms prior early, late, intermediate, and hybrid-early fusion variants. The largest margin appears on the survival-prediction task, while gains remain consistent on imaging and EHR tasks \emph{(e.g., up to $\approx$2.8 C-index points on BRCA, and consistent $\approx$0.37--1.4 point gains on imaging and $\approx$1.2--1.3 point gains on EHR in Table~\ref{tab:tab2}(c)).} These results support the core design choice behind CURE: a cascaded intermediate--late fusion pathway, augmented with multi-scale feature extraction and shared-information refinement, is crucial for capturing complementary cross-modal cues without sacrificing modality-specific structural information, thereby yielding robust shared representations and improved performance.

\medskip
\noindent\emph{\textbf{Module-wise ablations.}} We conduct module-wise ablations by decomposing the HyFuse layer into four modules, i.e., EMRC, HySAM, LLF, and SIR, showing that HyFuse’s performance gains are \emph{synergistic} rather than attributable to any single component (\textbf{Table~\ref{tab:tab3}(a)}). In particular, HySAM provides the primary performance lift over the backbone-only baseline, while EMRC and SIR yield consistent complementary gains and help stabilize shared-feature learning. Thus, it becomes clear that HySAM is the main contributor to performance improvements: removing HySAM from the full configuration causes the largest degradation (\emph{$\approx$2.4--9.9 points} across tasks), whereas removing SIR or EMRC incurs smaller but systematic drops (\emph{$\approx$0.4--1.6} and \emph{$\approx$0.3--1.2 points}, respectively). From an efficiency standpoint, EMRC is the key driver in CURE: adding EMRC to the HySAM, LLF, and SIR configuration nearly halves the parameters (14.7M$\rightarrow$7.71M) and reduces FLOPs (1.45$\rightarrow$0.59) \emph{while also improving performance}, confirming that CURE’s gains do not come from increased computational burden; hence, CURE balances the performance--efficiency trade-off through the coordinated design of these modules.

\medskip
\noindent\textbf{\emph{HySAM replacement ablations.}} We further ablate HySAM by replacing it with diverse prior attention baselines (\textbf{Table~\ref{tab:tab3}(b)}). Specifically, substituting HySAM with existing attention designs (e.g., CBAM \cite{woo2018cbam}, self-attention \cite{vaswani2017attention}, cross-attention \cite{chen2021crossvit}, QSAN \cite{shi2022qsan}) consistently reduces performance across all tasks, even under comparable computational costs. Among these replacements, the strongest alternative is MHDGA (the hyperbolic branch); however, the full HySAM pipeline still improves further by \emph{$\approx$0.8--1.7 points}, due to additionally leveraging the complementary MQIA branch and fusing both hyperbolic and quantum-informed cues within the same attention mixer. This substantiates the benefit of jointly exploiting hyperbolic hierarchy modeling and quantum-inspired interactions (with gated fusion) to learn more expressive shared representations, which in turn improves CURE performance.

%\noindent \textcolor{black}{\textbf{\emph{Comprehensive ablation studies}}} -- \emph{unimodal performance}, \textit{modality imbalance} in multimodal fusion, strategies for mitigating \textit{modality dominance}, and \emph{resolution-scaling evaluation} -- are \textcolor{black}{\textbf{\emph{summarized in App.~\ref{Ablation_Study}}}}.

%\vspace{-0.2cm}
%\section{Discussion.} 
%Our method learns modality-order-invariant robust shared features across heterogeneous modalities by preserving their structural properties and highlighting the benefit of effective and efficient cross-modal learning, and analyzing the associated computational complexity. \textcolor{purple}{\emph{Details appear in App.~\ref{discussion}}}.

%\vspace{-0.4cm}
%\subsection{Discussion} \label{Discussion}
%\vspace{-0.2cm}
%In \ref{discussion}, we show how our method learns richer complementary shared representations across heterogeneous modalities by preserving their structural properties. We further highlight the benefit of effective and efficient cross-modal learning and analyze the associated computational complexity \cite{an2025mlif}.  
%%%%%%%%%%%%%%%%%%%%%%%%%%%%%%%%%%%%%

%\vspace{-0.2cm} 
\section{Conclusion} \label{sec:concl}
%\vspace{-0.2cm} 
%We propose \texttt{MAIL}, which uses efficient multimodal cross attention module to unlock the full potential of shared representation learning across diverse modalities for classification and segmentation tasks in medical imaging. To ensure adversarial robustness, we introduce \texttt{Robust-MAIL}, incorporating \texttt{RPF} and \texttt{MAN}. Extensive evaluations demonstrate that our approach effectively utilizes the synergy across modalities and tasks, learning comprehensive, shared representations for multi-disease classification while ensuring adversarial robustness. Future work will explore advanced multiomics analysis and stronger adversarial defenses.
We present \texttt{CURE}, a novel, efficient, and effective multimodal fusion framework designed for analyzing heterogeneous medical data, making it ideal for resource-constrained, AI-driven healthcare settings. Evaluated across 16 diverse datasets, \texttt{CURE} achieves strong cross-modal generalization, outperforming leading state-of-the-art methods by up to 3.97\% in performance, with 85.7\% fewer parameters and 87.8\% lower \texttt{FLOPs}. Future work will focus on strengthening adversarial defenses and evaluating the framework on non-medical vision benchmarks to demonstrate adaptability across both medical and natural-vision domains.

\bibliographystyle{ACM-Reference-Format}
\small \bibliography{sample-base}

%\newpage
\appendix
%\newpage

%%%%%%%%%%

\section{Appendix -- Additional Related Work}\label{related}
\addcontentsline{toc}{section}{Appendix -- Additional  Related Works}

%%%%%%%
%\subsection{Research Gap Analysis}

We summarize the research gaps—highlighting how the \texttt{CURE} framework achieves an optimal performance–efficiency trade-off relative to existing multimodal fusion methods—in Table~\ref{tab:gap}.

%\section{Appendix -- \texttt{CURE} Algorithm} \label{algorithm}
%\addcontentsline{toc}{section}{Appendix D: CURE Algorithm}

%We present the pseudocode for the \texttt{CURE} framework in Algorithms \ref{al:al1}--\ref{al:al2}.

%%%%%%%%%

\section{Appendix -- Efficient Multimodal Residual Convolution Module} \label{emrc}
\addcontentsline{toc}{section}{Appendix E: Efficient Multimodal Residual Convolution Module}

%\subsection{\textbf{Efficient Multimodal Residual Convolution Module}} \label{EMRC}

%%%%%%%%%%%%%%%%%%%%%%%
%\vspace{-0.1cm}
\begin{figure*}[ht!]
    \centering
    \includegraphics[width=0.75\textwidth,
    %height=0.3\textheight
    ]{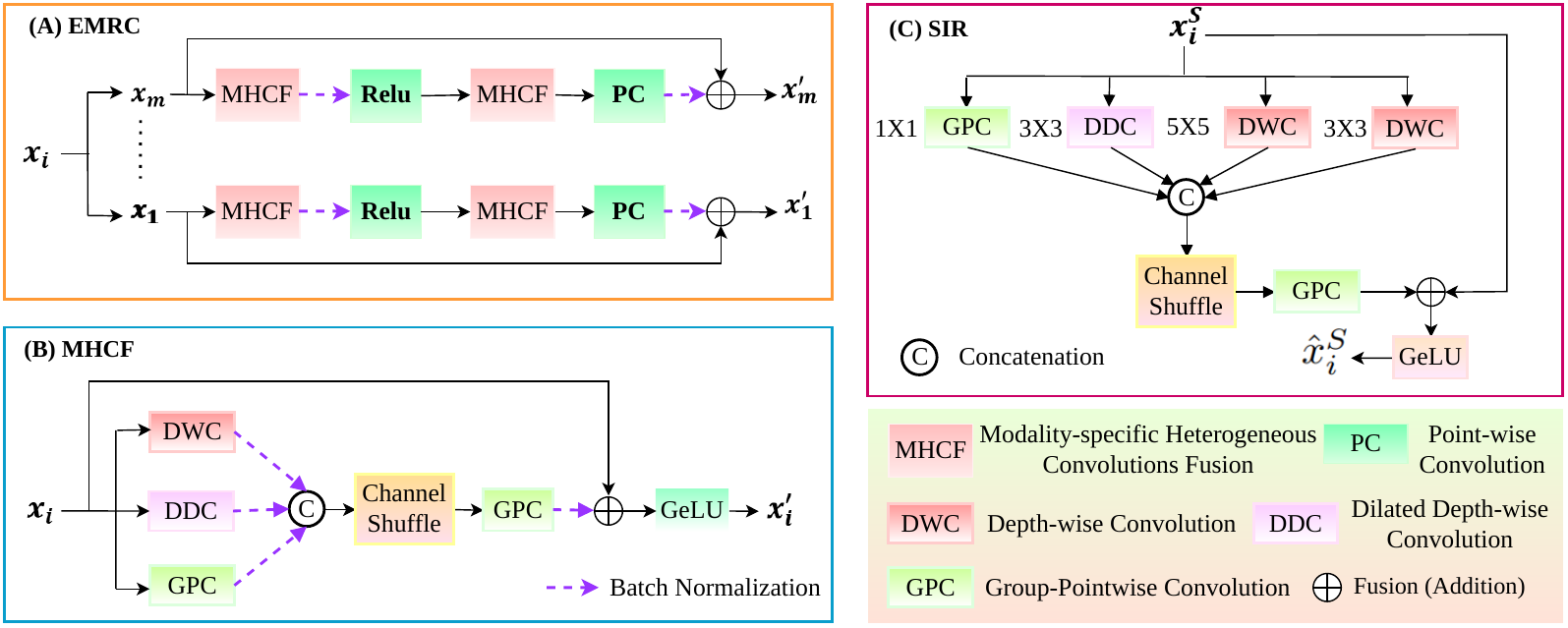}
    \vspace{-0.3cm}
    \caption{\small Architecture of the Efficient Multimodal Residual Convolution (\texttt{EMRC}) and Shared Information Refinement (\texttt{SIR}) modules. (A) The \texttt{EMRC} module integrates the Modality-specific Heterogeneous Convolutions Fusion (\texttt{MHCF}) block (shown in B) to progressively refine multimodal representations $x_i^{\prime}$, where $i\in[1:m]$. (C) The \texttt{SIR} module takes the shared representations $x_i^S$ produced by the \texttt{HySAM} module and refines them into $\hat{x}_i^S$ within each \texttt{HyFuse} layer, further enhancing representational diversity.} 
    \label{fig:fig6}
    \vspace{-0.0cm}
\end{figure*}

%%%%%%%%%%%%%%%%%%%%%%%%%%%%%%%%

We design the \textbf{E}fficient \textbf{M}ultimodal \textbf{R}esidual \textbf{C}onvolution (\texttt{EMRC}) module (ref. Fig. \ref{fig:fig6} (A)) to capture multi-scale spatial representations while ensuring low computational cost. The \texttt{EMRC} incorporates \textbf{M}odality-specific \textbf{H}eterogeneous \textbf{C}onvolutions \textbf{F}usion (\texttt{MHCF}) blocks (Fig. \ref{fig:fig6} (B)), along with BatchNorm ($\texttt{B}(\cdot)$), \texttt{ReLU} ($\mathcal{R}(\cdot)$), and pointwise convolution ($\texttt{PC}(\cdot)$) to facilitate progressive refinement. Specifically, \texttt{MHCF} uses diverse branch-wise convolutions -- Group-Pointwise (\texttt{GPC}), Dilated Depth-wise (\texttt{DDC}), and Depth-wise (\texttt{DWC}) -- at varying scales (\(1\!\times\!1\), \(3\!\times\!3\), \(5\!\times\!5\)), capturing heterogeneous spatial contexts and promoting branch-wise heterogeneity. The resulting contexts are fused and refined using a channel shuffle ($\texttt{CS}(\cdot)$) to facilitate inter-channel communication, thereby enhancing representational diversity, denoted as $x_{i}^{\prime} = \texttt{EMRC}(\cdot)$: 
%The \texttt{EMRC} is formally defined:
%\vspace{-0.5cm}
%\begin{footnotesize}
\begin{equation}  \small
\texttt{EMRC}(x_i) = x_i + \underbrace{\texttt{B}\Big(\texttt{PC}\big(\texttt{MHCF}(\mathcal{R}(\texttt{B}(\texttt{MHCF}(x_i))))\big)\Big)}_{\text{Progressive Refinement}},   
\label{eq:eq9_1}
\end{equation}

\begin{equation}  
\vspace{-0.2cm}
\small
  \texttt{MHCF}(x_i) = \texttt{CS} \bigg( \theta \Big( \underbrace{\forall_{sc \in \{1, 3, 5 \}} \texttt{HC}_{sc}(x_i)}_{\text{Branch-wise Heterogeneity}} \Big) \bigg) 
\label{eq:eq9_2}
\end{equation}
%\end{footnotesize}
%\vspace{-0.2cm}

\medskip
\noindent where the skip connection ensures stable gradient flow and regularization, $\texttt{HC}_{sc} \in \{\texttt{GPC}, \texttt{DDC}, \texttt{DWC} \}$, and $\theta$ denotes concatenation (fusion).

%%%%%%%%%%%

\section{Appendix -- Shared Information Refinement with Intermediate Fusion} \label{sir}
\addcontentsline{toc}{section}{Appendix F: Shared Information Refinement with Intermediate Fusion}

A detailed overview of the SIR module is illustrated in Fig.~\ref{fig:fig6} (C).

%%%%%%%%%%%%%%%

%\renewcommand{\thesection}{Appendix~G}

%%%%%%%%

%%%%%%%%%%%

% ============================================================
% Theoretical Analysis of CURE (drop-in for ICML 2-column style)
% ============================================================

%%%%%%%%%%%%

%%%%%%%%%%%%%%%%%%%%%%%%%%%

%%%%%%%%%%%%%%%%%%%%%%%%%%%%%%%%%%%%%%%%%%%%%%%%%%%%%%%

%\section{Appendix -- Proofs} \label{proofs}
%\addcontentsline{toc}{section}{Appendix C: Proofs}

%\renewcommand{\thesection}{Appendix~E}
\section{Appendix -- Data Augmentation} \label{data_augmentation}
\addcontentsline{toc}{section}{Appendix E: Data Augmentation}

For each training image \(x\) with label \(y\), we generate three
label-preserving variants: a \(20^\circ\) affine rotation, a 5-pixel
horizontal/vertical affine translation, and a \(3\times3\) Gaussian-blurred
image. These variants are assigned the same label \(y\), concatenated with
the original training set, and randomly shuffled, yielding a \(4\times\)
training set. This improves robustness to orientation, localization, and
acquisition-noise variations common in medical imaging.  

\begin{table}
%\vspace{-0.2cm}
  \setlength{\abovecaptionskip}{0pt}
  \captionsetup[table]{aboveskip=0pt, belowskip=0pt}
  \centering
  \caption{\small Characteristics of existing \texttt{MFL}  methods.}
  \label{tab:gap}
  \scalebox{0.9}{
    \setlength{\tabcolsep}{6pt}
    \begin{tabular}{@{}lcccc@{}}
      \toprule
      \textbf{Model} & \textbf{\makecell{Shared\\Representations}} & \textbf{\makecell{Cost-\\Effective}} & \textbf{\makecell{General-\\ization}} & \textbf{\makecell{Reli-\\ability}} \\
      \midrule
      %\texttt{POTTER}~\cite{zheng2023potter}          &      & \checkmark &       &       \\
      %\texttt{NAT}~\cite{hassani2023neighborhood}     &      & \checkmark &       &       \\
      %\texttt{DDA-Net}~\cite{cui2023dual}              &      & \checkmark &       &       \\
\rowcolor{gray!10}
      \texttt{Perceiver}~\citep{jaegle2021perceiver}   &   \checkmark   &            &       &       \\
      \texttt{GLORIA}~\citep{huang2021gloria}          & \checkmark &      &       &       \\
\rowcolor{gray!10}
      \texttt{HAMLET}~\citep{islam2020hamlet}          & \checkmark &      &       &       \\
      \texttt{MuMu}~\citep{islam2022mumu}              & \checkmark &      &       &       \\
\rowcolor{gray!10}
      $\texttt{M}^{3}\texttt{Att}$~\citep{liu2023multi} & \checkmark &      &       &       \\
      \texttt{MOTCAT}~\citep{xu2023multimodal}         & \checkmark &      &       &       \\
\rowcolor{gray!10}
      
      \texttt{DRIFA-Net}~\citep{dhar2024multimodal}              & \checkmark &      & \checkmark & \checkmark \\
      \texttt{HEALNet}~\citep{hemker2024healnet}       &      \checkmark   &  &       &       \\
      \rowcolor{orange!10}\textbf{\texttt{CURE} (Ours)}                               & \checkmark & \checkmark & \checkmark & \checkmark \\
      \bottomrule
    \end{tabular}
  }
  \vspace{-0.3cm}  
\end{table}

%%%%%%%%%%

\begin{table*}[htbp]
\centering
\caption{\small Performance comparison of \texttt{CURE-18} against \texttt{SOTA} methods on heterogeneous medical datasets: \texttt{BraTS-2021}, \texttt{SARS-CoV-2 CT-Scan}, \texttt{TCGA-GBMLGG}, \texttt{MHEALTH}, \texttt{CNMC-2019}, \texttt{Pneumonia}, and \texttt{UCI-HAR}, across multiple tasks. \textbf{Bold} indicates the best results.}
\vspace{-0.3cm}
\label{tab:tab4}

\resizebox{\textwidth}{!}{
\setlength{\tabcolsep}{4.5pt}
\begin{tabular}{l|cc|cc|c|cc|cc|cc|cc}
\toprule
\multirow{1}{*}{Datasets $\rightarrow$} 
& \multicolumn{2}{c|}{BraTS-2021} 
& \multicolumn{2}{c|}{SARS-CoV-2 CT-Scan} 
& \multicolumn{1}{c|}{TCGA-GBMLGG} 
& \multicolumn{2}{c|}{MHEALTH} 
& \multicolumn{2}{c|}{CNMC-2019} 
& \multicolumn{2}{c|}{Pneumonia} 
& \multicolumn{2}{c}{UCI-HAR} \\
\midrule
Models $\downarrow$ 
& ACC & AUC 
& ACC & AUC 
& C-Index 
& ACC & AUC 
& ACC & AUC 
& ACC & AUC 
& ACC & AUC \\
\midrule

\multirow{1}{*}{MFMSA} 
& 99.45{\scriptsize$\pm0.21$} & 99.30{\scriptsize$\pm0.56$}
& 98.82{\scriptsize$\pm0.81$} & 99.0{\scriptsize$\pm0.73$}
& 85.70{\scriptsize$\pm2.69$}
& 99.24{\scriptsize$\pm0.31$} & 99.10{\scriptsize$\pm0.45$}
& 96.70{\scriptsize$\pm1.19$} & 96.70{\scriptsize$\pm1.22$}
& 98.61{\scriptsize$\pm0.609$} & 98.48{\scriptsize$\pm0.384$}
& 96.30{\scriptsize$\pm0.93$} & 96.20{\scriptsize$\pm0.81$} \\
\rowcolor{gray!10}
\multirow{1}{*}{MSCAM} 
& 98.95{\scriptsize$\pm0.71$} & 98.90{\scriptsize$\pm0.64$}
& 98.50{\scriptsize$\pm0.36$} & 98.50{\scriptsize$\pm0.95$}
& 81.69{\scriptsize$\pm6.27$}
& 98.73{\scriptsize$\pm0.39$} & 98.85{\scriptsize$\pm0.52$}
& 96.21{\scriptsize$\pm1.32$} & 96.30{\scriptsize$\pm2.45$}
& 98.44{\scriptsize$\pm1.1$} & 98.27{\scriptsize$\pm0.95$}
& 96.18{\scriptsize$\pm0.88$} & 96.20{\scriptsize$\pm0.55$} \\
\midrule

\multirow{1}{*}{Perceiver} 
& 98.60{\scriptsize$\pm1.01$} & 98.80{\scriptsize$\pm0.89$}
& 98.36{\scriptsize$\pm1.24$} & 98.40{\scriptsize$\pm1.15$}
& 82.85{\scriptsize$\pm6.27$}
& 97.90{\scriptsize$\pm1.39$} & 98.25{\scriptsize$\pm1.23$}
& 93.50{\scriptsize$\pm3.65$} & 93.81{\scriptsize$\pm2.08$}
& 97.56{\scriptsize$\pm1.12$} & 98.0{\scriptsize$\pm0.90$}
& 95.0{\scriptsize$\pm3.771$} & 94.90{\scriptsize$\pm1.98$} \\
\rowcolor{gray!10}
\multirow{1}{*}{MuMu} 
& 99.0{\scriptsize$\pm0.74$} & 99.0{\scriptsize$\pm0.86$}
& 98.72{\scriptsize$\pm0.90$} & 98.54{\scriptsize$\pm1.02$}
& 82.63{\scriptsize$\pm5.87$}
& 98.56{\scriptsize$\pm1.14$} & 98.5{\scriptsize$\pm1.25$}
& 95.30{\scriptsize$\pm1.85$} & 95.30{\scriptsize$\pm2.15$}
& 98.10{\scriptsize$\pm0.65$} & 98.10{\scriptsize$\pm0.91$}
& 95.42{\scriptsize$\pm1.62$} & 95.60{\scriptsize$\pm1.98$} \\
\multirow{1}{*}{M$^3$Att} 
& 99.33{\scriptsize$\pm0.24$} & 99.30{\scriptsize$\pm0.37$}
& 99.15{\scriptsize$\pm0.54$} & 99.0{\scriptsize$\pm0.37$}
& 84.24{\scriptsize$\pm4.08$}
& 98.10{\scriptsize$\pm1.13$} & 98.25{\scriptsize$\pm1.38$}
& 95.52{\scriptsize$\pm2.40$} & 95.90{\scriptsize$\pm2.07$}
& 98.44{\scriptsize$\pm1.05$} & 98.50{\scriptsize$\pm0.97$}
& 95.37{\scriptsize$\pm1.41$} & 95.40{\scriptsize$\pm1.37$} \\
\rowcolor{gray!10}
\multirow{1}{*}{HEALNet} 
& 99.54{\scriptsize$\pm0.08$} & 99.60{\scriptsize$\pm0.1$}
& 98.90{\scriptsize$\pm0.93$} & 99.0{\scriptsize$\pm0.42$}
& 86.75{\scriptsize$\pm2.27$}
& 98.63{\scriptsize$\pm1.07$} & 98.80{\scriptsize$\pm0.90$}
& 96.12{\scriptsize$\pm1.05$} & 96.45{\scriptsize$\pm1.27$}
& 98.0{\scriptsize$\pm0.95$} & 98.0{\scriptsize$\pm1.34$}
& 95.52{\scriptsize$\pm1.84$} & 95.27{\scriptsize$\pm1.53$} \\
\multirow{1}{*}{DRIFA-Net} 
& 99.80{\scriptsize$\pm0.15$} & 99.80{\scriptsize$\pm0.1$}
& 99.23{\scriptsize$\pm0.42$} & 99.10{\scriptsize$\pm0.30$}
& 85.52{\scriptsize$\pm2.38$}
& 99.33{\scriptsize$\pm0.48$} & 99.0{\scriptsize$\pm0.48$}
& 96.85{\scriptsize$\pm1.12$} & 97.10{\scriptsize$\pm1.17$}
& 98.76{\scriptsize$\pm0.59$} & 98.44{\scriptsize$\pm0.61$}
& 96.30{\scriptsize$\pm0.85$} & 96.0{\scriptsize$\pm1.15$} \\
\midrule

\rowcolor{orange!10}
\multirow{1}{*}{\textbf{CURE-18}} 
& \textbf{100}{\scriptsize$\pm0.0$} & \textbf{100}{\scriptsize$\pm0.0$}
& \textbf{99.70}{\scriptsize$\pm0.15$} & \textbf{99.85}{\scriptsize$\pm0.10$}
& \textbf{89.34}{\scriptsize$\pm0.53$}
& \textbf{99.60}{\scriptsize$\pm0.37$} & \textbf{99.60}{\scriptsize$\pm0.25$}
& \textbf{97.54}{\scriptsize$\pm0.51$} & \textbf{97.83}{\scriptsize$\pm0.73$}
& \textbf{99.12}{\scriptsize$\pm0.31$} & \textbf{99.0}{\scriptsize$\pm0.202$}
& \textbf{96.95}{\scriptsize$\pm0.64$} & \textbf{97.10}{\scriptsize$\pm0.42$} \\
\bottomrule
\end{tabular}
}
\end{table*}

%%%%%%%%

% (Preamble) make sure you have:
% \usepackage{booktabs}
% \usepackage{graphicx} % for \resizebox
% Optional for row shading (only if you want it):
% \usepackage{colortbl}

\begin{table}[t]
\centering
\caption{\small Performance comparison of \texttt{CURE-18} against SOTA methods on two medical datasets: \texttt{KVASIR} and \texttt{MIT-BIH}, across multiple tasks. }
\label{tab:tab7}
\vspace{-0.3cm}
\small
\setlength{\tabcolsep}{3.5pt}
\renewcommand{\arraystretch}{1.1}

\resizebox{\columnwidth}{!}{%
\begin{tabular}{lccc|ccc}
\toprule
\textbf{Model} & \multicolumn{3}{c|}{\textbf{KVASIR}} & \multicolumn{3}{c}{\textbf{MIT-BIH}} \\
\cmidrule(lr){2-4}\cmidrule(lr){5-7}
& \textbf{ACC} & \textbf{F1} & \textbf{AUC} & \textbf{ACC} & \textbf{F1} & \textbf{AUC} \\
\midrule
MSCAM
& 89.52{\scriptsize$\pm3.58$} & 89.30{\scriptsize$\pm2.72$} & 89.61{\scriptsize$\pm5.08$}
& 99.27{\scriptsize$\pm0.25$} & 99.18{\scriptsize$\pm0.18$} & 99.50{\scriptsize$\pm0.05$} \\
\rowcolor{gray!10}
HEALNet
& 89.70{\scriptsize$\pm5.18$} & 89.45{\scriptsize$\pm4.71$} & 90.0{\scriptsize$\pm6.90$}
& 99.10{\scriptsize$\pm0.34$} & 99.10{\scriptsize$\pm0.25$} & 99.20{\scriptsize$\pm0.53$} \\
DRIFA-Net
& 91.47{\scriptsize$\pm4.12$} & 91.10{\scriptsize$\pm3.17$} & 91.60{\scriptsize$\pm4.59$}
& 99.60{\scriptsize$\pm0.15$} & 99.50{\scriptsize$\pm0.20$} & 99.60{\scriptsize$\pm0.15$} \\
\midrule
\rowcolor{orange!10}
\textbf{CURE-18}
& \textbf{94.74}{\scriptsize$\pm1.51$} & \textbf{94.30}{\scriptsize$\pm1.31$} & \textbf{94.52}{\scriptsize$\pm1.25$}
& \textbf{100}{\scriptsize$\pm0.0$} & \textbf{100}{\scriptsize$\pm0.0$} & \textbf{100}{\scriptsize$\pm0.0$} \\
\bottomrule
\end{tabular}%
}
\end{table}

\begin{table}[ht]
  \caption{\small Performance when swapping the modality order.}
  \label{tab:swap}
  \vspace{-0.3cm}
  \centering
  \scalebox{0.9}{
  \begin{tabular}{lcccc}
    \toprule
    Models & BRCA & MORT & HAM10000 & SIPaKMeD \\
    \midrule
    \rowcolor{gray!10}
    MOTCAT    & 56.05 & 90.38 & 95.87 & 93.24 \\
    DRIFA-Net & 55.63 & 91.20 & 98.41 & 95.58 \\
    \rowcolor{orange!10}
     \textbf{CURE-18}  & \textbf{63.12} & \textbf{93.45} & \textbf{99.60} & \textbf{96.35} \\
    \bottomrule
  \end{tabular}}
  \vspace{-0.3cm}
\end{table}

\section{Appendix -- Implementation Details} \label{implementataion_details}
\addcontentsline{toc}{section}{Appendix F: Implementation Details}

The evaluated baselines, \texttt{CURE} variants, metrics, losses, optimizer,
scheduler, and training hardware are summarized in
Sec.~\ref{experiments_analysis_results}. The sequential \texttt{CURE}/
\texttt{HyFuse} architecture is described in Secs.~\ref{proposed}--
\ref{sec_MSIL} and Fig.~\ref{fig:fig3}. We therefore report only the
additional adaptation and grouping details required for reproducibility.
For baselines originally designed for segmentation or other vision tasks,
we standardize the prediction interface by removing task-specific decoders
and heads, retaining the encoder/fusion trunk, applying global average
pooling to the final feature map, and attaching task-specific heads for
classification, mortality prediction, or survival estimation. Standalone
attention modules, e.g., \texttt{MFMSA} and \texttt{MSCAM}, are inserted into
their corresponding standard backbones, while their original decoders and
segmentation losses are discarded. Whenever applicable, each multimodal
baseline preserves its published fusion strategy, but all adapted models are
trained and evaluated using the same data splits, losses, metrics, and
optimization protocol (Sec.~\ref{experiments_analysis_results}).

\noindent\textbf{Training groups.}
To train on heterogeneous unpaired streams, we organize datasets into four
modality-basis groups and train one group at a time. For example, Group~1
uses \texttt{HAM10000} and \texttt{SIPaKMeD} imaging streams,
\texttt{TCGA-BRCA} multi-omics, and \texttt{MIMIC-III} mortality prediction;
Group~2 uses \texttt{PathMNIST} and \texttt{OrganAMNIST} imaging streams,
\texttt{TCGA-UCEC} multi-omics, and \texttt{MIMIC-III} ICD prediction. This
grouping is an implementation protocol only; the architecture itself follows
the general \texttt{MSIL} design in Sec.~\ref{sec_MSIL} and supports additional
available modalities without redesign. For four-stream groups, \texttt{CURE}
uses linear \(O(m)\) sequential fusion stages rather than \(O(m^2)\) all-pairs
fusion. Dropout with \(p=0.25\) is applied after fusion for regularization.

%%%%%%%

\section{Appendix -- Extensive Experimental Results} \label{Additional_Experiments}
\addcontentsline{toc}{section}{Appendix I: Extensive Experimental Results}

\subsection{Performance Comparison Analysis}

Table~\ref{tab:tab4} extends the evaluation to seven additional heterogeneous unpaired modalities spanning multi-modal MRI radiogenomic classification (\texttt{BraTS-2021}), CT screening (\texttt{SARS-CoV-2 CT-Scan}), multi-omics survival prediction (\texttt{TCGA-GBMLGG}), wearable time-series (\texttt{MHEALTH}, \texttt{UCI-HAR}), microscopy (\texttt{CNMC-2019}, and X-ray imaging \texttt{Pneumonia}). 
\texttt{CURE-18} ranks first
on all datasets. On classification tasks, it improves over the
strongest competing baseline by approximately \(0.2\)--\(0.9\) points,
while the largest gain appears on survival prediction, where it improves
\texttt{TCGA-GBMLGG} by \(+2.59\) C-index points. These results show
that \texttt{HyFuse} generalizes across imaging, time-series, and
multi-omics regimes without modality-specific redesign.

\medskip
\noindent We further test \texttt{CURE-18} in a limited two-stream setting using
\texttt{KVASIR} and \texttt{MIT-BIH} at \(224{\times}224{\times}3\)
resolution (Table~\ref{tab:tab7}). Even with only two modalities,
\texttt{CURE-18} consistently outperforms \texttt{MSCAM},
\texttt{HEALNet}, and \texttt{DRIFA-Net}, improving over the strongest
baseline by \(+3.27\) ACC points on \texttt{KVASIR} and reaching
ceiling performance on \texttt{MIT-BIH}. This indicates that the gains
of \texttt{CURE} are not contingent on using many modalities.

%%%%%%%%%%

\subsection{Input Order Robustness in Fusion}
\label{sec:input_order_robustness}

\texttt{CURE} is robust to input re-ordering across imaging
(e.g., \texttt{HAM10000}, \texttt{SIPaKMeD}), multi-omics
(e.g., \texttt{BRCA}), and clinical/\texttt{EHR} streams. As shown in
Table~\ref{tab:swap}, changing the modality order does not materially
degrade performance, since \texttt{HySAM} re-weights cross-modal
dependencies, \texttt{LLF}-gating suppresses missing or less informative
streams, and \texttt{SIR} refines the shared representation across cascaded
\texttt{HyFuse} layers.
Concretely, the first \texttt{HyFuse} layer fuses $m_1$ and $m_2$ to produce
$\{x_{i}^{s^{'}}\}_{i=1}^m$; the next layer fuses
$\{x_{i}^{s^{'}}\}_{i=1}^m$ with $m_3$, and the process continues until all
$m$ modalities are incorporated. Since each modality is first reshaped and
augmented to a common spatial size before \texttt{HyFuse}, \texttt{CURE}
does not require explicit temporal synchronization or spatial/structural
alignment, making it suitable for independently collected imaging, omics,
and \texttt{EHR} data.

%%%%%%%%

%%%%%%%%%%%%%

%%%%%%%%%

%\section{Appendix -- Qualitative Analysis} \label{Qualitative_Analysis}
%\addcontentsline{toc}{section}{Appendix K: Qualitative Analysis}

%%%%%%%%%

\end{document}